\documentclass{article}

\usepackage{microtype}
\usepackage{graphicx}
\usepackage{booktabs} 
\usepackage{caption}
\usepackage{subcaption}

\usepackage{amsmath}
\usepackage{mathtools}
\usepackage{amssymb}
\usepackage{hyperref}
\usepackage{lipsum}
\usepackage{multirow}
\usepackage{mathtools}
\usepackage[nameinlink,capitalise]{cleveref}
\usepackage{lscape}
\usepackage{bibentry}

\newcommand{\tth}[1]{^{(#1)}}
\def\[#1\]{\begin{align}#1\end{align}}
\DeclareMathOperator*{\argmax}{argmax}

\usepackage[acronym,smallcaps,nowarn,section,nogroupskip,nonumberlist]{glossaries}
\glsdisablehyper{}
\newacronym{eot}{eot}{Expectation Over Time}
\newacronym{mcmc}{mcmc}{Markov-Chain Monte-Carlo}
\newacronym{pgd}{pgd}{Projective Gradient Descent}
\newacronym{bpda}{bpda}{Backward Pass Differentiable Approximation}
\newacronym{dpp}{dpp}{Denoising Projective Purification}
\newacronym{adp}{adp}{Adaptive Denoising Purification}
\newacronym{ebm}{ebm}{Energy-Based Model}
\newacronym{sm}{sm}{Score-Matching}
\newacronym{dsm}{dsm}{Denoising Score-Matching}
\newacronym{ncsn}{ncsn}{Noise Conditional Score Network}
\newacronym{spsa}{spsa}{Simultaneous Perturbation Stochastic Approximation}
\newacronym{dct}{dct}{Discrete Cosine Transform}
\newacronym{bim}{bim}{Basic Iterative Method}
\newcommand{\dee}{\mathrm{d}}
\newcommand{\R}{\mathbb{R}}
\newcommand{\E}{\mathbb{E}}
\newcommand{\norm}[1]{\left\Vert {#1} \right\Vert}
\newcommand{\algorithmicbreak}{\textbf{break}}
\newcommand{\BREAK}{\STATE \algorithmicbreak}

\usepackage{hyperref}

\usepackage[accepted]{icml2021}
    
\icmltitlerunning{Adversarial Purification with Score-based Generative Models}

\begin{document}

\twocolumn[
\icmltitle{Adversarial Purification with Score-based Generative Models}



\icmlsetsymbol{equal}{*}

\begin{icmlauthorlist}
\icmlauthor{Jongmin Yoon}{ka}
\icmlauthor{Sung Ju Hwang}{ka,ait}
\icmlauthor{Juho Lee}{ka,ait}
\end{icmlauthorlist}

\icmlaffiliation{ka}{Korea Advanced Institute of Science and Technology, Daejeon, Korea}
\icmlaffiliation{ait}{AITRICS, Seoul, Korea}

\icmlcorrespondingauthor{Jongmin Yoon}{jm.yoon@kaist.ac.kr}
\icmlcorrespondingauthor{Juho Lee}{juholee@kaist.ac.kr}

\icmlkeywords{adversarial purification, score-based generative model, energy-based model}

\vskip 0.3in
]



\printAffiliationsAndNotice{}  

\begin{abstract}
While adversarial training is considered as a standard defense method against adversarial attacks for image classifiers, adversarial purification, which purifies attacked images into clean images with a standalone purification model, has shown promises as an alternative defense method. Recently, an \gls{ebm} trained with \gls{mcmc} has been highlighted as a purification model, where an attacked image is purified by running a long Markov-chain using the gradients of the \gls{ebm}. Yet, the practicality of the adversarial purification using an \gls{ebm} remains questionable because the number of \gls{mcmc} steps required for such purification is too large. In this paper, we propose a novel adversarial purification method based on an \gls{ebm} trained with \gls{dsm}. We show that an \gls{ebm} trained with \gls{dsm} can quickly purify attacked images within a few steps. We further introduce a simple yet effective randomized purification scheme that injects random noises into images before purification. This process screens the adversarial perturbations imposed on images by the random noises and brings the images to the regime where the \gls{ebm} can denoise well. We show that our purification method is robust against various attacks and demonstrate its state-of-the-art performances.
\end{abstract}
\glsresetall
\section{Introduction}
\label{sec:introduction}
Image classifiers built with deep neural networks are known to be vulnerable to adversarial attacks, where an image containing a small perturbation imperceptible to human completely changes the prediction results~\citep{DBLP:journals/corr/GoodfellowSS14, DBLP:conf/iclr/KurakinGB17}. There are various methods that aim to make classifiers robust to such adversarial attacks, and \emph{adversarial training} \citep{madry2018towards, DBLP:conf/icml/ZhangYJXGJ19}, in which a classifier is trained with adversarial examples, is considered as a standard defense method due to its effectiveness.

Another approach for the adversarial defense is to \emph{purify} attacked images before feeding them to classifiers. This strategy, referred to as \emph{adversarial purification}~\citep{srinivasan2019robustifying, pmlr-v97-yang19e, shi2021online}, learns a \emph{purification model} whose goal is to remove any existing adversarial noise from potentially attacked images into clean images so that they could be correctly classified when fed to the classifier. The purification model is usually trained independently of the classifier and does not necessarily require the class labels for training; thus, it is less likely to affect the clean image classification accuracy compared to the adversarial training methods. The most common way is for adversarial purification to learn a \emph{generative model} over the images~\citep{samangouei2018defensegan, song2018pixeldefend, Schott2019Towards, Ghosh2019Resisting} such that one can restore clean images from attacked images.

Recently, along with the advances in learning \glspl{ebm} with deep neural networks, adversarial purification methods using an \gls{ebm} trained with \gls{mcmc} as a purification model have been highlighted~\citep{DuIGEBM, Grathwohl2020Your, hill2020stochastic}. Hinging on the memoryless behavior of \gls{mcmc}, these methods purify attacked images by running a large number of sampling steps defined by Langevin dynamics. When started from an attacked image, in the long run, Langevin sampling will eventually bring the attacked image to a clean image that is likely to be generated from the data distribution. However, the success of the purification heavily depends on the number of sampling steps, and unfortunately, the number of steps required for stably purify attacked images is too large to be practical. 

In this paper, we propose a novel adversarial purification method using an \gls{ebm} trained with \gls{dsm}~\citep{JMLR:v6:hyvarinen05a,Song2019NCSN} as a purification model. Unlike an \gls{ebm} trained with \gls{mcmc} estimating the energy function, \gls{dsm}~\citep{Vincent2010dae} learns the \emph{score function} that can \emph{denoise} the perturbed samples, which is more closely related to the adversarial purification because the purification can be thought as denoising of the adversarial attacks. 
We show that an \gls{ebm} trained with \gls{dsm}, using the \emph{deterministic update scheme} that we propose, can quickly purify the attacked images within several orders of magnitude fewer steps than the previous methods. We further propose a simple technique to enhance the robustness of our purification method by injecting noises to images before the purification. The intuition behind this is, by injecting noises relatively larger than adversarial perturbations, we can make the adversarial perturbations negligible and also at the same time convert the images to the familiar noisy images that are similar to the ones seen during the training with \gls{dsm}. As our model is facilitated by random noises around the images, the classifier with randomly purified images can be interpreted as a \emph{randomized smoothing classifier}~\citep{Cohen2019randomized, Pinot2020Randomization}. Since the noise distribution over clean images and attacked images overlap, we validate that our denoising framework delivers \emph{certified robustness} over any threat models, implying that the attacked image is guaranteed to be provably robust regardless of the attacks driven.

We verify our method on various adversarial attack benchmarks and demonstrate its state-of-the-art performance. Unlike the previous purification methods, we validate our defense method using a strong attack that approximates the gradients of the entire purification procedure. We show the state-of-the-art performance of our methods on several datasets using various attacks.

\section{Backgrounds}
\label{sec:backgrounds}

\subsection{Adversarial training, preprocessing, and purification}
Consider a classifier neural network $g_\phi: \R^D \to \R^K$. A neural network robust to adversarial attacks can be trained with the \emph{worst-case risk minimization},
\begin{equation}\label{eq:advloss}
\min_{\phi} \mathbb{E}_{p_\text{data}(x,y)}\left[\max_{x'\in\mathcal{B}(x)}\mathcal{L} \left( g_{\phi} \left( x' \right) , y \right)\right],
\end{equation}
where $\mathcal{L}$ is a loss function and $\mathcal{B}\left(x\right)$ is the threat model at $x$ (i.e., the domain of all attacked images from $x$). The exact evaluation and optimization of this objective are infeasible, so one must resort to an approximation. A predominant approach is \emph{adversarial training}~\citep{madry2018towards, DBLP:conf/icml/ZhangYJXGJ19, Carmon2019RST},
which optimizes a surrogate loss computed from restricted classes of adversarial examples during training. For instance, \citet{madry2018towards} approximates the inner maximization loop in \eqref{eq:advloss} with the average loss computed for the chosen attacked images.
\begin{align}\label{eq:madryloss}
\min_{\phi} \mathbb{E}_{p_\text{data}(x,y)}\left[\E_{x'\in\mathcal{A}(x)}\mathcal{L} \left( g_{\phi} \left( x' \right) , y \right)\right],
\end{align}
where $\mathcal{A}(x_i)$ is the set of images obtained by attacking $x$ with chosen classes of attacks. Due to the relaxation of the full threat model, the resulting network may still be vulnerable to attacks not included in $\mathcal{A}(x)$.

Another approach for adversarial defense is \emph{preprocessing}, 
where input images are preprocessed with auxiliary transformations before classification.
Let $f_{\theta}$ be a preprocessor transforming images. The training objective is then
\begin{align}
\min_{\phi,\theta} \E_{p_\text{data}(x,y)}\left[
\max_{x' \in \mathcal{B}(x)} \mathcal{L}(g_\phi(f_\theta(x')), y)
\right].
\end{align}
The maximum over the threat model $\mathcal{B}(x)$ may be approximated with an average over known classes of adversarial attacks or average over stochastically transformed inputs. Such transformations include adding stochasticity into the input images or adding discrete or non-differentiable transforms into the input images~\citep{Guo18Countering, Dhillon2018SAP, Buckman2018Thermometer, Xiao2020Enhancing}, making the gradient estimation with respect to the loss function $\nabla_x \mathcal{L}(g_\phi(f_\theta(x)), y)$ harder for an attacker.

Finally, in \emph{adversarial purification}, a \emph{generative model} that can restore clean images from attacked images is additionally trained and used as a preprocessor~\citep{samangouei2018defensegan, song2018pixeldefend, srinivasan2019robustifying, Ghosh2019Resisting, hill2020stochastic, shi2021online}, where the preprocessor $f_\theta$ corresponds to the purification process defined with the generative model. 

\subsection{Energy-based models and adversarial purification}\label{sec:ap}
An \gls{ebm}~\citep{Lecun2006ebm} defined on $\R^D$ is a probabilistic model  whose density function is written as
\[
     p_\theta(x) = \frac{\exp(-E_\theta(x))}{Z_\theta},
\]
where $E_\theta(x): \R^D \to \R$ is the \emph{energy function} and $Z_\theta = \int_x \exp(-E_\theta(x))\dee x$ is the normalization constant. Since $E_\theta(x)$ is not subject to any constraint (e.g., integrate to one), an \gls{ebm} provides great flexibility in choosing the form of the model. However, due to the intractable normalization constant, computing the density and learning the parameter $\theta$ requires approximation. Roughly speaking, there are three methods to learn an \gls{ebm}; maximum-likelihood with \gls{mcmc}, score matching~\citep{JMLR:v6:hyvarinen05a}, and noise-contrastive estimation~\citep{Gutmann2010nce}. 

Training an \gls{ebm} with maximum likelihood involves the computation of the gradient
\[
\lefteqn{\nabla_\theta\E_{p_\text{data}(x)}[\log p_\theta(x)]}\nonumber\\
&= \E_{p_\theta(x)}[\nabla_\theta E_\theta(x)] - \E_{p_\text{data}(x)}[\nabla_\theta E_\theta(x)].
\]
Here, the first term evaluates the intractable expectation over the model distribution $p_\theta(x)$ and is usually approximated with Monte-Carlo approximation. Recently, drawing samples from $p_\theta(x)$ from a Markov-chain defined with Langevin dynamics
has been demonstrated to work well, even for high-dimensional energy functions constructed with deep neural networks~\citep{Nijikamp2019bebm,DuIGEBM}. Using Langevin sampling, one can simulate $p_\theta(x)$ by running a Markov chain,
\[\label{eq:ebm_sgld}
 x_t = x_{t-1} - \frac{\alpha}{2} \nabla_x E_\theta(x_{t-1}) + \sqrt{\alpha}\varepsilon,
\]
where $\alpha > 0$ is a step size, $x_0$ is a randomly initialized starting point, and $\varepsilon \sim \mathcal{N}(0, I)$. This can be used not only for the training but also for generating novel samples from trained \gls{ebm}s. Based on this, \citet{DuIGEBM,Grathwohl2020Your} showed that an \gls{ebm} trained with \gls{mcmc} can purify adversarially attacked images by running Langevin sampling starting from the attacked images. \citet{hill2020stochastic} further developed this by adjusting the optimization process of an \gls{ebm} to make it better convergent for long \gls{mcmc} chains, and demonstrated that the adversarial purification with long-run Langevin sampling can successfully purify attacked images into clean images. The key for the successful purification is the ``long-run" \gls{mcmc} sampling. The number of \gls{mcmc} steps required for purification is typically more than 1,000, which is costly even with modern GPUs.

\subsection{Denoising score matching}
\gls{sm}~\citep{JMLR:v6:hyvarinen05a} is a density estimation technique that learns the \textit{score function} of the target density instead of directly learning the density itself. Let $p_\text{data}(x)$ be a target density defined on $\R^D$, and $\log p_\theta(x)$ be a model. The score model $\nabla_x \log p_\theta(x) := s_\theta(x): \R^D \to \R^D$ is then trained to approximate the true score function $\nabla_x \log p_\text{data}(x)$ by minimizing the objective $\E_{p_\text{data}(x)}[\frac{1}{2}\norm{s_\theta(x)-\nabla_x\log p_\text{data}(x)}]$, and under mild conditions, this can be shown to be equivalent to minimizing
\begin{align}\label{eq:sm_objective}
    \E_{p_\text{data}(x)}\left[\frac{1}{2}\norm{s_\theta(x)}^2 + \mathrm{tr}(\nabla_x s_\theta(x)) \right].
\end{align}
\citet{JMLR:v6:hyvarinen05a} showed that this objective gives a consistent estimator of the true parameter $\theta$. \gls{sm} is useful for training an \gls{ebm} because computing the score functions does not require computing the intractable normalizing constants. However, still, the basic version of score matching with objective \eqref{eq:sm_objective} does not scale to high-dimensional data due to the term $\mathrm{tr}(\nabla_x s_\theta(x))$.

To avoid computing the term $\mathrm{tr}(\nabla_x s_\theta(x))$, \gls{dsm}~\citep{DBLP:journals/neco/Vincent11} slightly tweaks the objective \eqref{eq:sm_objective}. The basic idea of \gls{dsm} is to learn the score function of the \emph{perturbed} data. Given a pre-specified noise distribution $q(\tilde{x}|x)$, \gls{dsm} minimizes the following objective,
\begin{align}\label{eq:dsm_objective}
    \E_{q(\tilde x|x)p_\text{data}(x)}\left[ \frac{1}{2}\norm{s_\theta(\tilde x) - \nabla_{\tilde x}\log q(\tilde x| x)}^2 \right].
\end{align}
This modified objective is well-defined, provided that the noise distribution is smooth, and if the noise is small so that $q(x) := \int q(x|x') p_\text{data}(x')\dee x' \approx p_\text{data}(x)$, \gls{dsm} finds the same solution as the original \gls{sm}. A common choice for $q(\tilde x | x)$ is the Gaussian distribution centered at $x$, $q(\tilde x|x) = \mathcal{N}(\tilde x ; x, \sigma^2 I)$. In such case, \eqref{eq:dsm_objective} reduces to
\begin{align}\label{eq:dsm_denoising}
    \ell(\theta,\sigma) = \E_{q(\tilde x|x) p_\text{data}(x)}\left[
    \frac{1}{2\sigma^4} \norm{
    \tilde x + \sigma^2s_\theta(\tilde{x}) - x
    }^2
    \right].
\end{align}
That is, $s_\theta(\tilde x)$ is trained to recover the original data $x$ from the perturbed data $\tilde x$, as in denoising autoencoders~\citep{Vincent2008dae,Vincent2010dae}. 
\section{Adversarial purification with score-based generative models}
\label{sec:methods}

\begin{figure}[t]
    \centering
    \includegraphics[width=0.95\linewidth]{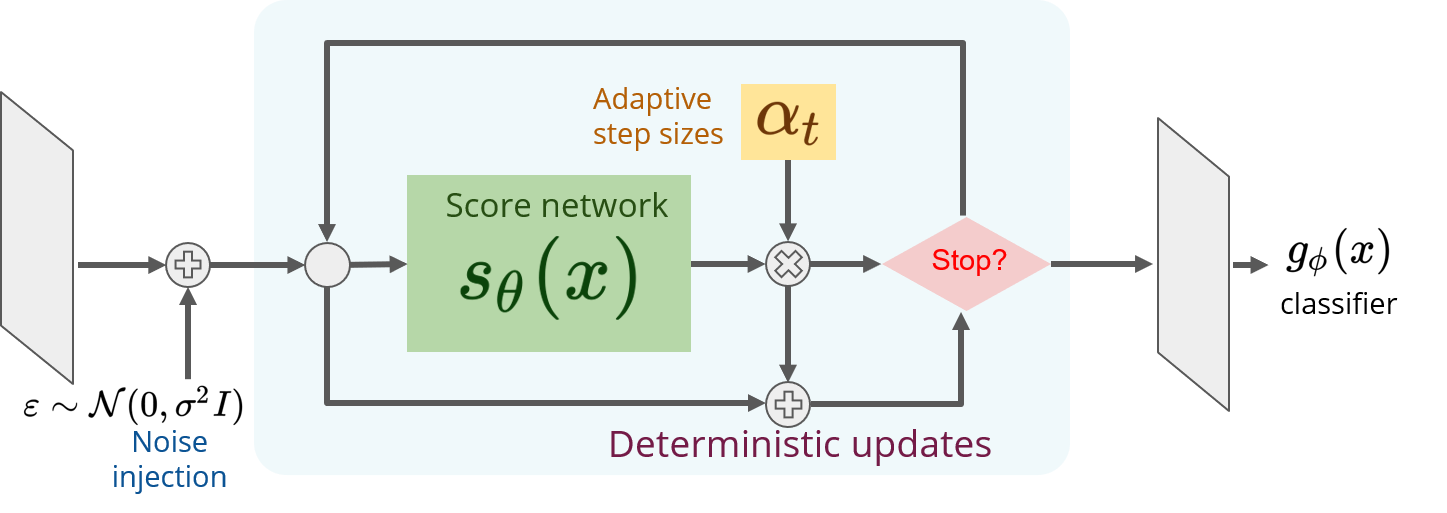}
    \caption{A conceptual diagram describing our approach.}
    \label{fig:concept}
\end{figure}

\subsection{Denoising score matching for adversarial purification}
While previous adversarial purification methods using \glspl{ebm} employ maximum likelihood training with \gls{mcmc}, we propose to use \gls{dsm} to train an \gls{ebm} to be used for purification. As we discussed above, \gls{dsm} aims to learn a score network $s_\theta(x)$ pointing the direction of restoring original samples from perturbed samples. Considering that the adversarial purification can be understood as a denoising procedure, we conjecture that \gls{dsm} is a better option to train an \gls{ebm} as it is learned with the objective more directly related to the adversarial purification. The maximum likelihood training aims to learn the energy function $E_\theta(x)$, and the gradients $\nabla_x E_\theta(x)$ is obtained as a byproduct. Hence, unless trained perfectly, accurately predicting the energy function does not necessarily mean accurately predicting the gradients $\nabla_x E_\theta(x)$ that are actually used for purification.

In this work, we adopt the recently proposed \gls{ncsn}~\citep{Song2019NCSN, Song2020NCSNv2} for our model. An \gls{ncsn} is trained with a modified \gls{dsm} incorporating multiple noise levels, where inputs are perturbed with multiple noise levels $\{\sigma_j\}_{j=1}^L$ instead of a single noise $\sigma$. The training objective based on this multi-scale noise is 
\begin{equation}\label{eq:ncsn_loss}
\mathcal{L}(\theta, \{\sigma_j\}_{j=1}^L) = \sum_{j=1}^L \sigma_j^2 \ell(\theta, \sigma_j),
\end{equation}
where $\ell(\theta, \sigma_j)$ is the \gls{dsm} objective \eqref{eq:dsm_denoising} with a specific noise level. We choose the noise levels $\{\sigma_j\}_{j=1}^L$ following the guidance in \citet{Song2020NCSNv2}. Training with multiple noises exposes the score network $s_\theta(x)$ to the various perturbed samples, and this may be advantageous for the adversarial purification of various attacked images.

\citet{Song2020NCSNv2} observed that the norm of the noise-conditioned score function trained with \cref{eq:ncsn_loss} is approximately reciprocal to the noise level, i.e., $\norm{s_\theta(x, \sigma)}_2 \propto 1/\sigma$, and proposed to use the conditional score function as $s_\theta(x, \sigma) = s_\theta(x)/\sigma$. Throughout the paper, we follow this parameterization for our purification model.

\subsection{Purification by deterministic updates}
\label{subsec:deterministic}
Let $s_\theta(x)$ be a trained score network and $g_\phi$ be a classifier network with softmax output. Given an attacked image $x'$, previous adversarial purification methods would run a stochastic Markov chain driven by Langevin dynamics, with $s_\theta(x)$ in place of $-\nabla_x E_\theta(x)$. Instead, we propose to \emph{deterministically update} the samples with the learned scores. That is, starting from $x_0 = x'$, execute the following updates for $t \geq 1$,
\[
x_t = x_{t-1} + \alpha_{t-1} s_\theta(x_{t-1}),
\]
where $\{\alpha_t\}_{t \geq 0}$ are step-sizes that may be tuned with a validation set or chosen adaptively using the algorithm that we will describe in short. 
\begin{figure}
    \centering
    \includegraphics[width=0.7\linewidth]{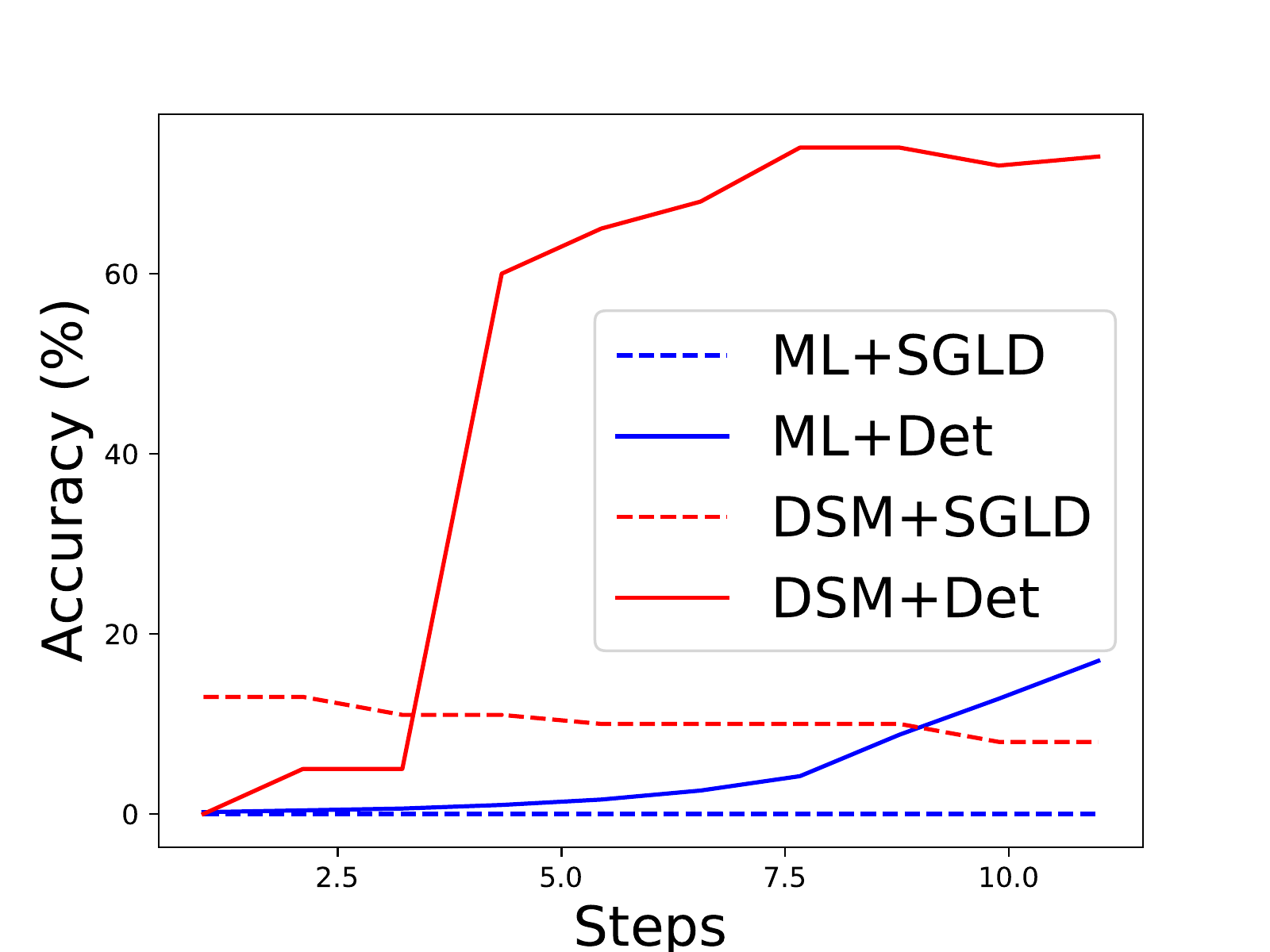}
    \caption{The accuracy against the \gls{bpda} attack on CIFAR10. ML denotes the maximum likelihood training with \gls{mcmc}, and Det denotes deterministic updates.}
    \label{fig:det_vs_sgld}
\end{figure}

While this deterministic update does not guarantee convergence, we empirically found that it improves classification accuracy much faster than stochastic updates. We also found that the deterministic update also slightly improves the speed of purification 
for \gls{ebm} trained with \gls{mcmc} in the short run, but \gls{dsm} works much better with deterministic updates~(\cref{fig:det_vs_sgld}).

After running $T$ steps of deterministic updates, we can pass $x_T$ through the classifier to get a prediction.
We may stop the iteration before $T$ steps when the adaptive step size $\alpha_t$ becomes less then the pre-specified threshold $\tau$.


\subsection{Random noise injection before purification}
As we will show in \cref{sec:experiments}, the defense method based on the deterministic purification described in \cref{subsec:deterministic} can successfully defend most of the adversarial attacks, but it is vulnerable to the strong attack based on the gradient estimation of the full purification process. We propose a simple enhancement to our algorithm that can even defend this strong attack and improve classification accuracy. The idea is simple; We add a random Gaussian noises to images before purification. Given a potentially attacked image $x'$, the purification proceeds as
\[
x_0 &= x' + \varepsilon, \quad \varepsilon \sim \mathcal{N}(0, \sigma^2I) \nonumber\\
x_t &= x_{t-1} + \alpha_{t-1} s_\theta(x_{t-1}).\label{eq:rand_purification}
\]
The intuition behind this is as follows. Assume that an attacked image $x'$ contains an adversarial perturbation as $x' = x + \nu$, and we add a Gaussian noise $\varepsilon \sim \mathcal{N}(0, \sigma^2I)$ to form $x_0 = x + \nu + \varepsilon$. Since the norm of $\nu$ is bounded due to the perceptual indistinguishability constraint, the added noise $\varepsilon$ can ``screen out’’ the relatively small perturbation $\nu$. Also, recall that the score network $s_\theta(x)$ is trained to denoise images perturbed by Gaussian noises. Adding Gaussian noises makes $x_0$ more similar to the data used to train the score network.

The initial noise level $\sigma$ is an hyperparameter to be specified. Following the popular heuristic used for kernel methods~\citep{garreau2018large}, we choose it as the median of the pairwise Euclidean distances divided by the square root of the input dimension $\sqrt{D}$. The initial noise level selected with this heuristic is $\sigma=0.25$ for both CIFAR-10 and CIFAR-100.


Due to the randomness of the purification induced by noise injection, we execute the multiple purification runs and take the ensemble as
\[
\hat y = \argmax_{k\in[K]} \frac{1}{S} \sum_{s=1}^S [g_\phi( x_T\tth{s})]_k,
\]
where $\{x\tth{s}_t\}_{t=1}^T$ is an instance of a purification using \eqref{eq:rand_purification}. Note that the computation for multiple runs of purification (along with $s$ indices) can be parallelized.

Injecting noises before purification now turns our deterministic update updates into a random purification method, and thus the classifier taking randomly purified image can be interpreted as a randomized smoothing classifier~\citep{Cohen2019randomized,Pinot2020Randomization}. In \cref{sec:experiments}, we actually show that the randomized smoothing classifier derived from our method has certified robustness over any norm-bounded threat models.

\subsection{Adaptive step sizes}\label{sec:lr}
The step-sizes $\{\alpha_t\}_{t \geq 0}$ are important hyperparameters that can affect the purification performance. While these can be tuned with additional vaildation set~\footnote{When tuning these values on validation set, we choose a single learning rate $\alpha_t = \alpha$ for all $t \geq 0$ for feasibility of search.}, we propose a simple yet effective adaptation scheme that can choose proper step-sizes during the purification.

Let $x_t$ be an intermediate point during purification. If trained properly, in a small local neighbor of $x_t$, there exists $\sigma_t$ such that $s_\theta(x,\sigma_t)$ is close to the score function of a Gaussian distribution $\mathcal{N}(\mu, \sigma_t^2I)$ for some $\mu$ and $\sigma_t$. That is,
\[
s_\theta(x_t) \approx - \frac{x_t-\mu}{\sigma_t}.
\]
Let $\alpha_t$ be a step size for $x_t$. We want the score of the updated point $x_{t+1} = x_t + \alpha_t s_\theta(x_t)$ to have a decreased scale with ratio $(1-\lambda)$ for some $\lambda \in (0,1)$, since the score decreases to zero as we move $x_t$ closer to the local optimum ($\mu)$.
\[
s_\theta(x_{t+1}) = (1-\lambda) s_\theta(x_t),
\]
and this leads to
\[
\alpha_t s_\theta(x_t) \approx \lambda \sigma_t s_\theta(x_t) \Rightarrow \alpha_t = \lambda \sigma_t.
\]
Now, to estimate $\sigma_t$, we move $x_t$ along the direction of $s_\theta(x)$ by a small step-size $\delta$ to compute $x' = x_t + \delta s_\theta(x_t)$. Then $\sigma_t$ can be approximated as
\[
s_\theta(x') - s_\theta(x_t) &\approx \frac{-\delta s_\theta(x_t)}{\sigma_t} \nonumber\\
\Rightarrow \sigma_t &\approx \frac{\delta \norm{s_\theta(x_t)}^2}{\norm{s_\theta(x_t)}^2 - s_\theta(x')^\top s_\theta(x_t)}.
\]
Hence, we get
\[
\alpha_t &= \lambda\delta \left(1 - \frac{s_\theta(x_t)^\top s_\theta(x')}{\norm{s_\theta(x_t)}^2} \right)^{-1},
\]
as our step size. We find this adaptive learning rates works well without much tuning of the parameters $\lambda$. For all experiments, we fixed $\lambda = 0.05$ and $\delta = 10^{-5}$.

We call our adversarial purification method combining all the ingredients (\gls{dsm} + deterministic updates + noise injection + adaptive step sizes) \gls{adp}. The purification procedure with \gls{adp} is summarized in \cref{alg:adp}.

\begin{algorithm}[tb]
\small
   \caption{Adversarial purification with \gls{adp}}
   \label{alg:adp}
\begin{algorithmic}
   \STATE {\bfseries Input:} an input $x$, the score network $s_\theta$, the classifier $g_\phi$, noise scale $\sigma$, number of purification runs $S$, number of steps per each purification run $T$, adaptive learning rate parameters $\lambda$ and $\delta$, purification stopping threshold $\tau$.
   \FOR{$s=1$ {\bfseries to} $S$}
    \STATE $x\tth{s}_0 \leftarrow x + \varepsilon, \quad \varepsilon \sim \mathcal{N}(0, \sigma^2I)$.
    \FOR{$t=1$ {\bfseries to} $T$}
    \STATE $x' \leftarrow x\tth{s}_{t-1} + \delta s_\theta(x_{t-1}\tth{s})$.
    \STATE $\alpha_{t-1} \leftarrow \lambda \delta\bigg(1 - \frac{s_\theta(x_t\tth{s})^\top s_\theta(x')}{\norm{s_\theta(x_t\tth{s})}^2}\bigg)^{-1}$
    \STATE $x\tth{s}_t \leftarrow x\tth{s}_{t-1} + \alpha_{t-1}s_\theta(x\tth{s}_{t-1})$.
    
    \IF {$\norm{s_\theta(x\tth{s}_t)} < \tau$}
      \STATE $x_{T}\tth{s}\gets x_{t}\tth{S}$.
      \BREAK
    \ENDIF
    
   \ENDFOR
   \ENDFOR
   \STATE $\hat{y} = \argmax_k \frac{1}{S}\sum_{s=1}^S  [g_\phi(x_{T}\tth{s})]_k$.
   \STATE {\bfseries Return }$\hat y$.
\end{algorithmic}
\end{algorithm}

\section{Related Works}
\label{sec:related}
\paragraph{Adversarial training}
\citet{DBLP:journals/corr/SzegedyZSBEGF13, DBLP:journals/corr/GoodfellowSS14} discovered that visually imperceptible signals for humans could effectively disturb the neural network image classifiers. Adversarial training~\citep{DBLP:conf/iclr/KurakinGB17, madry2018towards} learns a robust classifier by augmenting those adversarial examples at the training phase, and has been shown to be the most reliable defense method. Some techniques such as regularization \citep{DBLP:conf/icml/ZhangYJXGJ19,Wang2020Improving} or self-supervised learning \citep{Carmon2019RST} further improves robustness performance. 

\paragraph{Preprocessing for adversarial defense}
Many existing works propose adversarial defense by preprocessing attacked images via auxiliary transformations or stochasticity before classification. They include thermometer encoding \citep{Buckman2018Thermometer}, stochastic activation pruning \citep{Dhillon2018SAP}, image quilting \citep{Guo18Countering}, matrix estimation \citep{pmlr-v97-yang19e} and discontinuous activation \citep{Xiao2020Enhancing}. Due to the transformations, those methods cause the phenomenon so as called \emph{obfuscated gradients} that makes estimating gradients for gradient-based attack difficult, such as shattered gradients or vanishing/exploding gradients. However, \citet{pmlr-v80-athalye18a, Tramer2020} designed strong attacks that can bring down the robust accuracy of those defense methods to almost zero.

\paragraph{Adversarial purification methods}
There also have been various adversarial purification methods. \citet{samangouei2018defensegan} proposed defense-GAN which trains a generator restoring clean images from attacked images. \citet{song2018pixeldefend} showed that an autoregressive generative model can detect and purify adversarial examples. \citet{srinivasan2019robustifying} proposed a purification method with Metropolis-Adjusted Langevin algorithm (MALA) applied to denoising autoencoders~\citep{Vincent2008dae}. \citet{Grathwohl2020Your, DuIGEBM} showed the promise of \glspl{ebm} trained with \gls{mcmc} can purify adversarial examples, and \citet{hill2020stochastic} demonstrated that long-run \gls{mcmc} with \glspl{ebm} can robustly purify adversarial examples.
\section{Experiments}
\label{sec:experiments}
\glsunset{pgd}
\begin{table*}[t]
\centering
\caption{List of attacks considered. After each update, the output is projected with $x_{i+1}=\prod_{\mathcal{B}_{\infty}(x_0,b)} x_{i+1} '$. Here $f_{\theta}:\mathbb{R}^D\to\mathbb{R}^D$ is the full purification model, $s_{\theta}:\mathbb{R}^D\to\mathbb{R}^D$ is the score network that consists the purification and $g_{\phi}:\mathbb{R}^D\to\mathbb{R}^K$ is the classifier, where $D$ is the dimension of data and $K$ is the number of classes. For \gls{spsa} attack, $v_i$ is uniformly sampled from $\{-1,1\}^D$. For all of our experiments, we fix $\alpha_i=2/255$ and $\varepsilon=0.5$.}
\footnotesize
\begin{tabular}{lll}
\toprule
Attack name & Type & Updating rule to derive $x_{i+1}'$ \\ 
\midrule
Full gradient & White-box & $x_i + \alpha_i \mathtt{sign}\nabla_x \mathcal{L} \left( \left(g_{\phi}\circ f_{\theta}\right) \left( x \right) , y \right)\rvert_{x=x_i}$ \\
Classifier \gls{pgd} & Preprocessor-blind & $x_i + \alpha_i \mathtt{sign}\nabla_x \mathcal{L} \left( g_{\phi}\left( x \right) , y \right)\rvert_{x=x_i}$ \\
\gls{bpda} \cite{pmlr-v80-athalye18a} & Adaptive & $x_i + \alpha_i \mathtt{sign}\nabla_x \mathcal{L} \left( g_{\phi}\left( x \right) , y \right)\rvert_{x=f_{\theta}\left(x_i\right)}$ \\
Joint attack (score) & Adaptive & $x_i +\alpha_i \left(\varepsilon\mathtt{sign} (s_{\theta}(x_i)) +(1-\varepsilon)\mathtt{sign}(\nabla_x \mathcal{L}(g_{\phi}(x),y)\rvert_{x=x_i}\right)$ \\
Joint attack (full) & Adaptive & $x_i +\alpha_i \left(\varepsilon\mathtt{sign} (f_{\theta}(x_i)-x_i) +(1-\varepsilon)\mathtt{sign}\nabla_x \mathcal{L}(g_{\phi}(x),y)\rvert_{x=x_i}\right)$ \\
\gls{spsa} \cite{pmlr-v80-uesato18a} & Black-box & $x_i + \alpha_i\mathtt{sign} \sum_{j=1}^N \frac{\mathcal{L}\left(\left(\left( g_{\phi}\circ f_{\theta} \right) \left(x+\varepsilon v_j\right), y \right) - \mathcal{L}\left(\left( g_{\phi}\circ f_{\theta} \right) \left(x-\varepsilon v_j\right), y \right)\right)\cdot v_j}{2N\varepsilon}$ \\
\bottomrule
\end{tabular}
\label{table:attacklist}
\end{table*}
\glsreset{pgd}

In this section, we validate our defense method \gls{adp} from various perspectives. First, we evaluate \gls{adp} under strongest existing attacks on $\ell_\infty$-bounded threat models and compare it to other state-of-the-art defense methods including adversarial training and adversarial purification. Then we show the certified robustness of our model on $\ell_2$-bounded threat models and compare it to other existing randomized classifiers. 
We further verify the perceptual robustness of our method with common corruptions~\citep{hendrycks2018benchmarking} on CIFAR-10. We further validate ours on a variety of datasets including MNIST, FashionMNIST, and CIFAR-100.



For all experiments, we use WideResNet~\citep{Zagoruyko2016WRN} with depth 28 and width factor 10, having 36.5M parameters. For the score model, we use \gls{ncsn} having 29.7M parameters. For the purification methods including ours, we use naturally-trained classifier, i.e., we do not use adversarial training or other augmentations. Unless otherwise stated, for \gls{adp}, we fixed the adaptive step size parameters $(\lambda, \delta) = (0.05, 10^{-5})$, and computed ensembles over $S=10$ purification runs, i.e., we take 10 random noise injection over Gaussian distribution $\varepsilon\sim\mathcal{N}(0,\sigma^2 I)$, followed by clipping to $[0,1]$. We fixed the purification stopping threshold $\tau$ is given by $0.001$. As aforementioned, the noise standard deviation $\sigma$ was fixed to $0.25$ for all experiments, and please refer to the supplementary material for the results with different values of $\sigma$. As an ablation, we also tested \gls{adp} without noise injection $(\sigma=0.0)$. We found that purification with adaptive step sizes does not work well without the noise injection, so used manually tuned step-size schedule using validation sets. Please refer to the supplementary material for detailed model description and settings. For all of the attacks described later, we fix the threat model to an $\ell_2$ $\varepsilon$-ball with $\varepsilon=8/255$.

\subsection{List of adversarial attacks}\label{sec:attacks}
The full list of attacks we considered is shown in \cref{table:attacklist} with types and updating rules. We briefly describe the attacks in detail.
\paragraph{Preprocessor-blind attack} This is the weakest adversarial attack on the list, where an attacker has full access to the classifier but has no access to the purification model. Attacks under this scenario are sometimes considered as \emph{gray-box} attacks in literature, but we consider this as a transfer-based black-box attack~\citep{pmlr-v80-uesato18a} with source model $g_{\phi}$ and target model $g_{\phi}\circ f_{\theta}$. We test with the \gls{pgd} attack on the classifier $g_\phi$.

\paragraph{Strong adaptive attack}
Our purification algorithm consists of multiple iterations through neural networks, so might cause obfuscated gradient problems. Hence, we also validate our defense method with strong adaptive attacks, including \gls{bpda} \citep{pmlr-v80-athalye18a} and its variants. 
We test the basic version of \gls{bpda} where the purification process $f_\theta(x)$ is approximated with identity function. We also test the following modifications of the \gls{bpda} customized to the adversarial purification methods.
\begin{itemize}
    \item Joint attack (score): updates the input by weighted sum of the classifier gradient and score network output. If an attacked image has low score norm, then it will not be purified by our algorithm.
    \item Joint attack (full): updates the input by weighted sum of the classifier gradient and difference between an original input and purified input.
\end{itemize}

Since our defense method contains random noise injection, we validate our defense method with \gls{eot}~\citep{pmlr-v80-athalye18a} attacks together with strong adaptive attacks.

\paragraph{Score-based black-box attack}
Even when an attacker does not have access to a model and its gradient with respect to a loss function, the gradient can still be estimated with large number of samples. One of such approaches is \gls{spsa}~\citep{Spall1992SPSA,pmlr-v80-uesato18a}, where the random samples near an input are drawn and the approximate gradient is obtained by expected value of gradients approximated with the finite-difference method.
One caveat of these attacks is that the number of samples required to estimate gradients can be large. In our setting, we set the number of queries to 1,280 to make the attack strong enough.

\paragraph{Common corruption}
\citet{hendrycks2018benchmarking} proposed \emph{common corruptions}, a class of 75 frequently occuring visual perturbations and suggested to use them for testing robustness of classifiers. We test our defense method on CIFAR-10-C where those common corruptions are applied to CIFAR-10 dataset. While ours and other adversarial defense methods are not designed to defend these perturbations, they can still be a good way to test robustness of defense methods.

\subsubsection{Evaluation: preprocessor-blind attacks}
\begin{table}
\caption{CIFAR-10 results for Preprocessor-blind attacks. The \gls{pgd} attacks to the classifier is performed at $\ell_{\infty}$ $\varepsilon$-ball with  $\varepsilon=8/255$. The results borrowed from the references are marked with $*$.}
\vspace{0.05in}
\scriptsize
\centering
\setlength{\tabcolsep}{5pt}
\begin{tabular}{llll}
\toprule
\multirow{2}{*}{Models} & \multicolumn{2}{c}{Accuracy (\%)} & \multirow{2}{*}{Architecture}\\
  & Standard & Robust &  \\
\midrule
Raw WideResNet & 95.80 & 0.00 & WRN-28-10 \\
\gls{adp} ($\sigma=0.1$) & 93.09 & 85.45 & WRN-28-10 \\
\gls{adp} ($\sigma=0.25$) & 86.14 & 80.24 & WRN-28-10 \\
\midrule
\multicolumn{3}{l}{Adversarial purification methods}\\
\cite{hill2020stochastic} & 84.12 & 78.91 & WRN-28-10 \\
\cite{shi2021online}* & 96.93 & 63.10 & WRN-28-10 \\
\cite{DuIGEBM}* & 48.7 & 37.5 & WRN-28-10 \\
\cite{Grathwohl2020Your}* & 75.5 & 23.8 & WRN-28-10 \\
\cite{pmlr-v97-yang19e}* & & & \\
\hspace{3mm}$p=0.8\to1.0$ & 94.9 & 82.5 & ResNet-18 \\
\hspace{3mm}$p=0.6\to0.8$ & 92.1 & 80.3 & ResNet-18 \\
\hspace{3mm}$p=0.4\to0.6$ & 89.2 & 77.4 & ResNet-18 \\
\cite{song2018pixeldefend}* & & & \\
\hspace{3mm}Natural + PixelCNN & 82 & 61 & ResNet-62 \\
\hspace{3mm}AT + PixelCNN & 90 & 70 & ResNet-62 \\
\midrule
\multicolumn{3}{l}{Adversarial training methods, transfer-based}\\
\cite{madry2018towards}* & 87.3 & 70.2 & ResNet-56 \\
\cite{DBLP:conf/icml/ZhangYJXGJ19}* & 84.9 & 72.2 & ResNet-56 \\
\bottomrule
\end{tabular}
\label{table:pb}
\end{table} 

The evaluation of defense methods on CIFAR-10 with preprocessor-blind attacks are shown in \cref{table:pb}. 
We present the evaluation results of \gls{adp} for preprocessor-blind attacks on CIFAR-10 dataset, and compare them with other preprocessor-based methods and adversarial training methods.  Except for~\citep{shi2021online}, all the methods do not have any knowledge about the attacks during training. 

Since the preprocessor-blind attacks are considered as a special case of \emph{transfer-based black-box attacks}, we also compare ours
to adversarial training methods tested with transfer-based black-box attacks~\citep{madry2018towards,DBLP:conf/icml/ZhangYJXGJ19}. The transfer-based black-box attacks assume that an attacker can access the training data, and thus can train a substitute model generating adversarial examples with them. The results for the transfer based attacks are borrowed from \citet{Dong2020Benchmark}.

We observe that \gls{adp} successfully purifies attacked images and shows high robust-accuracy on preprocessor-blind attacks while maintaining high natural accuracy. 

\subsubsection{Evaluation: strong adaptive attacks}
\begin{table}[t]
\scriptsize
\centering
\caption{CIFAR-10 results for adaptive attacks at $\ell_\infty$ $\varepsilon$-ball with $\varepsilon=8/255$. We compare our proposed method with other recently proposed preprocessor-based defense methods, and adversarial training methods with white-box attacks for reference. The results borrowed from the references are marked with $*$.}
\vspace{0.05in}
\setlength{\tabcolsep}{5pt}
\begin{tabular}{lllll}
\toprule
Models & \multicolumn{2}{c}{Accuracy (\%)} & \multirow{2}{*}{Architecture} \\
\hspace{3mm}Attacks & Natural & Robust & \\
\midrule
\gls{adp} ($\sigma=0.25$) & 86.14 &  &  \\
\hspace{3mm}\gls{bpda} 40+\gls{eot} &  & \textbf{70.01} & WRN-28-10 \\
\hspace{3mm}\gls{bpda} 100+\gls{eot} &  & \textbf{69.71} & WRN-28-10 \\
\hspace{3mm}Joint (score)+\gls{eot} &  & 70.61 & WRN-28-10 \\
\hspace{3mm}Joint (full)+\gls{eot} &  & 78.39 & WRN-28-10 \\
\hspace{3mm}\gls{spsa} &  & 80.80 & WRN-28-10 \\
\midrule
\gls{adp} ($\sigma=0.0$) & 90.60 & & \\
\hspace{3mm}\gls{bpda} &  & 76.87 & WRN-28-10 \\
\hspace{3mm}Joint (score) &  & 80.81 & WRN-28-10 \\
\hspace{3mm}Joint (full) &  & 80.58 & WRN-28-10 \\
\hspace{3mm}\gls{spsa} &  & 47.6 & WRN-28-10 \\
\midrule
\multicolumn{4}{l}{Adversarial purification methods} \\
\cite{hill2020stochastic} & & & \\
\hspace{3mm}\gls{bpda} 50+\gls{eot} & 84.12 & 54.90 & WRN-28-10 \\
\cite{song2018pixeldefend}* & & & \\
\hspace{3mm}\gls{bpda} & 95.00 & 9 & ResNet-62 \\
\cite{pmlr-v97-yang19e}* & & &\\
\hspace{3mm}\gls{bpda} 1000 & 94.8 & 40.8 & ResNet-18 \\
\hspace{4mm}(+AT, $p=0.4\to0.6$) & 88.7 & 55.1 & WRN-28-10 \\
\hspace{4mm}(+AT, $p=0.6\to0.8$) & 91.0 & 52.9 & WRN-28-10 \\
\hspace{3mm}Approx. Input & 89.4 & 41.5 & ResNet-18 \\
\hspace{3mm}Approx. Input (+AT) & 88.7 & 62.5 & ResNet-18 \\
\cite{shi2021online}* & & & \\
\hspace{3mm} Classifier \gls{pgd} 20 & 91.89 & 53.58 & WRN-28-10    \\ 
\midrule
\multicolumn{4}{l}{Adversarial training methods} \\
\hspace{3mm}\cite{madry2018towards}* & 87.3 & 45.8 & ResNet-18 \\
\hspace{3mm}\cite{DBLP:conf/icml/ZhangYJXGJ19}*  & 84.90 & 56.43 & ResNet-18 \\
\hspace{3mm}\cite{Carmon2019RST} & 89.67 & 63.1 & WRN-28-10 \\
\hspace{3mm}\cite{gowal2020uncovering}* & 89.48 & 64.08 & WRN-28-10 \\
\bottomrule
\end{tabular}
\label{table:ad}
\end{table}
We present our evaluation results for strong adaptive attacks on CIFAR-10 dataset as introduced in \cref{sec:attacks}. \cref{table:ad} shows the evaluation results for various adaptive attacks. For \gls{bpda} and its variants, we assume that an attacker knows the exact step sizes used for the purification, and the attacks are designed with them. For \gls{spsa} attacks, we use 1,280 batch size to make the attack strong enough. For \gls{bpda}+\gls{eot} attack and its variants, we take $15$ different noisy inputs and get expected loss over them for each attack step. We provide further results with different number of \gls{eot} attacks in~\cref{sec:bpdaeot}.

Ours generally shows the best results, successfully defending most of the strong adaptive attacks. One can see that, the random noise injection ($\sigma=0.25)$ generally improves the robust accuracy compared to the ones for without random noise injection $(\sigma=0.0)$. Especially, without random noise injection, ours fails on \gls{spsa} while it doesn't with random noise injection.


\begin{figure}[t]
    \centering
    \includegraphics[width=0.6\linewidth]{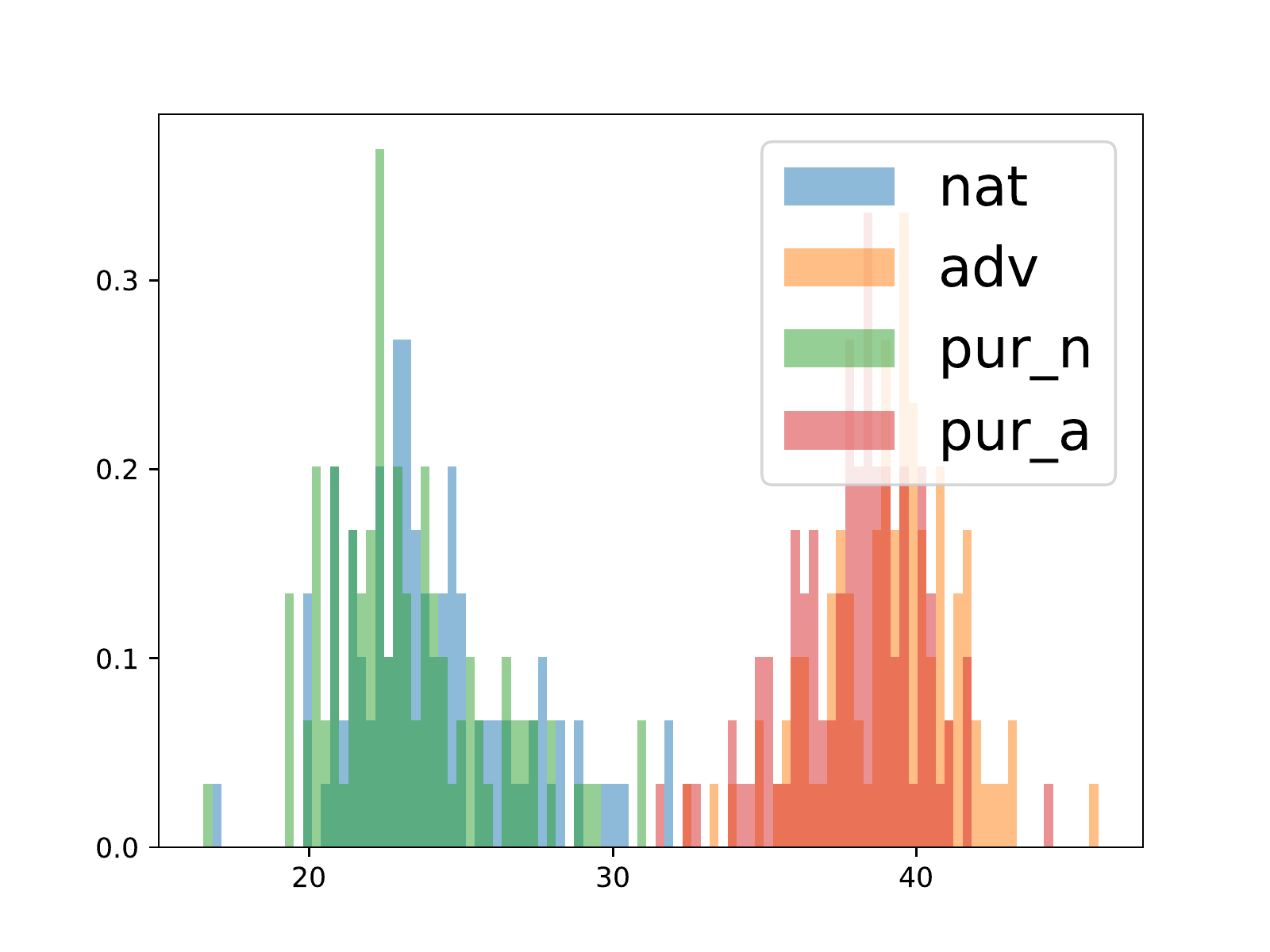}
    \caption{Histogram of score norms $\norm{s_\theta(x)}_2$ for natural, adversarial, and purified images.
    Pur\_a (pur\_n) denotes score norms of one-step purified adversarial (natural) images. The x-axis and y-axis represent the score norm $\norm{s_{\theta}(x)}_2$ and the density, respectively.}
    \label{fig:scorehistogram}
\end{figure}

\subsubsection{Evaluation: certified robustness}
\begin{figure}
\captionsetup{justification=raggedright, singlelinecheck=off}
\begin{subfigure}{.49\linewidth}
\includegraphics[width=\linewidth]{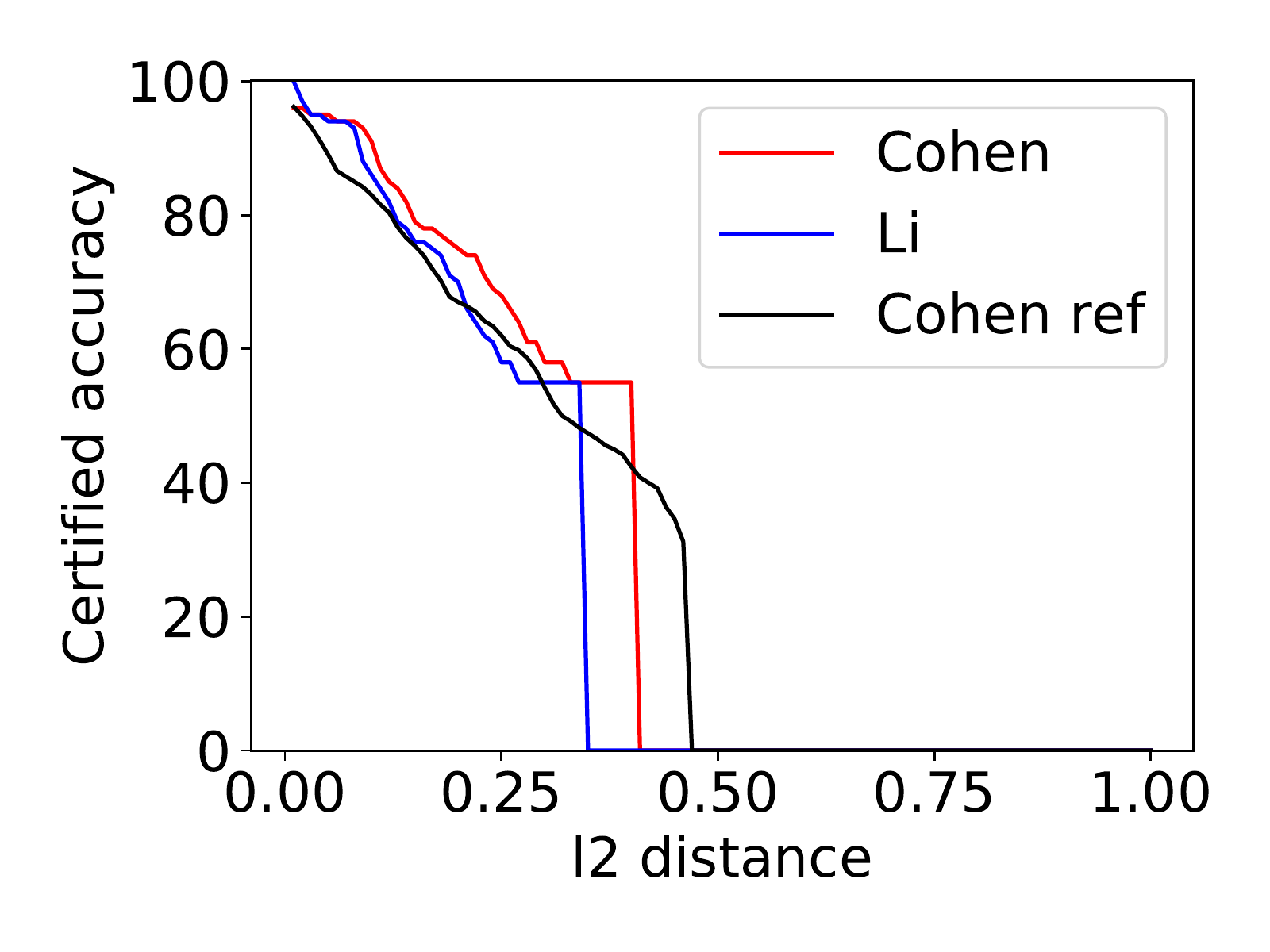}
\end{subfigure}
\begin{subfigure}{.49\linewidth}
\includegraphics[width=\linewidth]{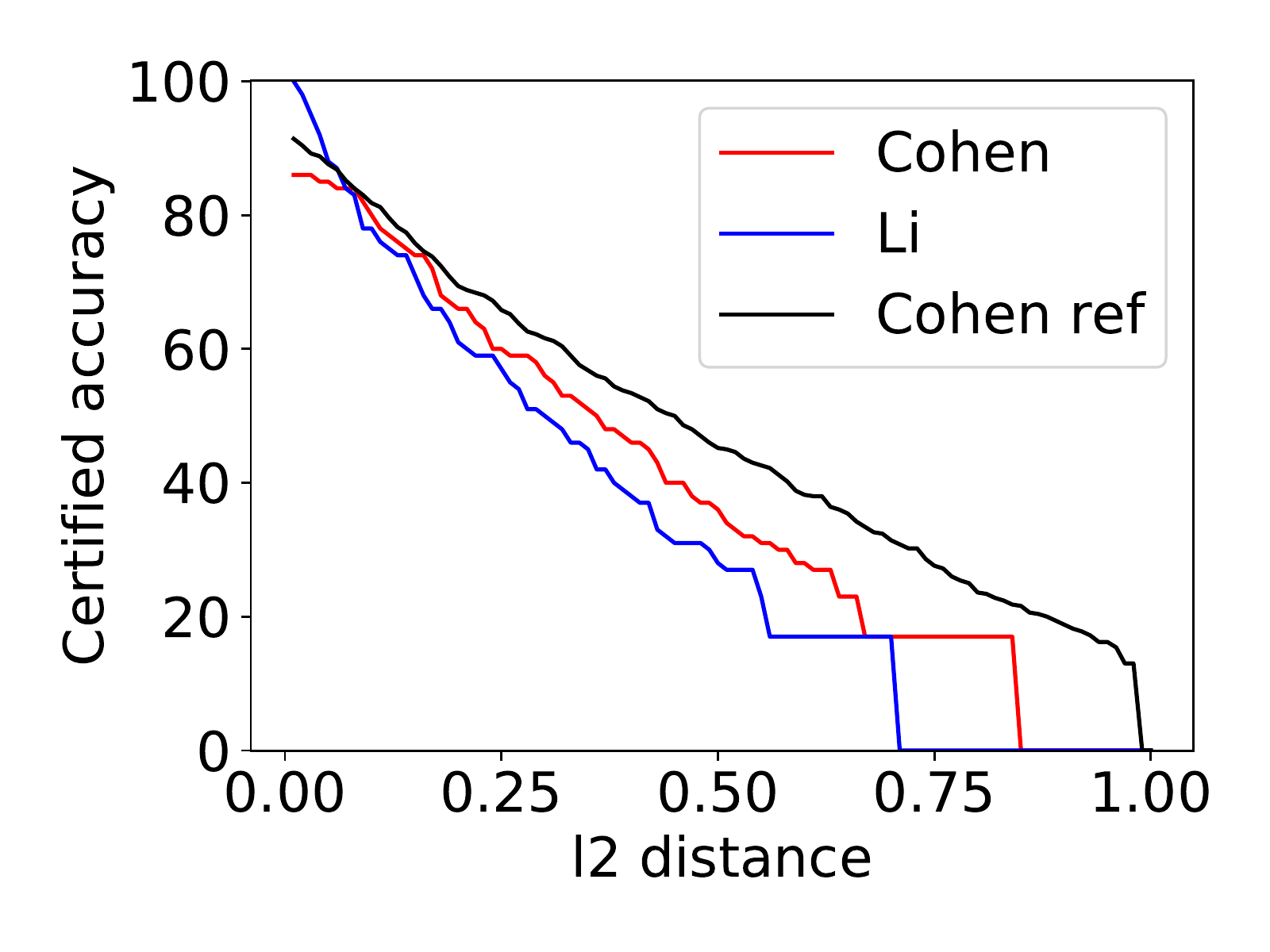}
\end{subfigure}
\caption{Approximate certified accuracy estimated by randomized smoothing. Left: $\sigma=0.12$, Right: $\sigma=0.25$, where $\sigma$ is the noise level. The {\color{red}red} and {\color{blue}blue} lines are the approximate certified accuracy of our model with respect to measures given from \citet{Cohen2019randomized} and \citet{Li2018Second}, respectively. The black line is the reference certified accuracy of randomized smoothing classifier from \citet{Cohen2019randomized}.} 
\label{fig:certified}
\end{figure}
A classifier is \emph{certifiably robust} within an area including an input $x$ if the classifier predicts a constant label inside the area. A randomized smoothing classifier $h_{\phi}$ from the base classifier network $g_{\phi}$ is given by
$h_{\phi}(x)=\argmax_{k\in[K]}\,[\mathbb{E}_{\varepsilon}[g_{\phi}(x+\varepsilon)]]_k$, where $\varepsilon \sim \mathcal{N}(0, \sigma^2I)$. According to~\citep{Cohen2019randomized}, the certified radius of the randomized smoothing classifier is defined by
\begin{align}\label{eq:certified}
R&=\frac{\sigma}{2}(\Phi^{-1}(p_A)-\Phi^{-1}(p_B)),
\end{align}
where $p_A$ and $p_B$ are the probabilities of top-2 probable labels after Gaussian noise $\mathcal{N}(0,\sigma^2 I)$ is injected, and $\Phi^{-1}$ denotes the inverse standard Gaussian CDF. We report our approximate certified accuracy in \cref{fig:certified}, by sampling noisy images for 100 times per image on 100 sampled images at the test set in CIFAR-10 dataset, and compare this to \citet{Cohen2019randomized} evaluated for 500 sampled images. We show that we outperform previous results at noise level $\sigma=0.12$, and show comparable certified accuracy at $\sigma=0.25$.

\subsubsection{Evaluation: Common corruptions}
\begin{table}
\caption{CIFAR-10-C results for common corruption. The results borrowed from the references are marked with $*$.}
\centering
\scriptsize
\setlength{\tabcolsep}{3pt}
\begin{tabular}{llllll}
\toprule
 & Accuracy & Noise & Blur & Weather & Digital \\
\midrule
Raw WideResNet   & 71.89 & 42.37 & 72.74 & 86.23 & 78.83\\
\gls{adp} ($\sigma=0.25$) & 77.45 & 84.68 & 75.52 & 77.98 & 73.42 \\
\gls{adp} ($\sigma=0.25$)+Detection & 78.96 & 84.57 & 71.76 & 84.34 & 76.60 \\
\gls{adp} ($\sigma=0.1$) & 76.25 & 86.88 & 68.55 & 78.88 & 73.36 \\ 
\gls{adp}, ($\sigma=0.0$)  & 80.49 & 84.58 & 73.12 & 86.39 & 78.89 \\
\gls{adp}, ($\sigma=0.0$ + DCT) & 80.74 & 84.96 & 73.11 & 87.26 & 78.69\\
\gls{adp}, ($\sigma=0.0$ + AugMix) & 82.63 & 87.52 & 75.20 & 88.82 & 80.20\\
\gls{adp}, ($\sigma=0.0$ + DCT+AugMix) & 82.40 & 85.05 & 75.80 & 88.11 & {81.30}\\
\midrule
\multicolumn{5}{l}{Adversarial training methods}\\
\cite{DBLP:conf/icml/ZhangYJXGJ19} & 75.63 & 77.83 & 78.37 & 74.98 & 71.88\\
\cite{Carmon2019RST} & 80.40 & 81.20 & {83.44} & 80.19 & 76.98\\
\cite{Cohen2019randomized} & 73.70 & 81.48 & 72.60 & 72.19 & 70.46 \\
\midrule
\multicolumn{5}{l}{Training classifiers with augmentations}\\
\cite{Hendrycks2020augmix}* & 88.78 & 84.66 & 89.68 & 90.93 & 88.47\\
\cite{wang2021tent}* & 89.52 & 85.61 & 89.68 & 91.88 & 89.95\\
\cite{hossain2020robust}* & 89.17 & 86.80 & 88.75 & 91.25 & 89.28\\
\bottomrule
\end{tabular}
\label{table:cc}
\end{table}
We present the evaluation results on CIFAR-10-C. The results are summarized in \cref{table:cc}. Please refer to \cref{sec:cifar10cfull} for full results. We found that \gls{adp} with noise injection underperforms \gls{adp} without noise injection. Unlike the adversarial perturbations whose norms are bounded, the norms of the corruptions applied to images are not bounded, so they are not effectively screened by Gaussian noise injection. Still, the average performance with or without noise injection surpasses other adversarial training baselines (note that \citet{Carmon2019RST} uses extra unlabeled dataset so is not directly comparable). We could further improve the robust accuracy by exploiting \gls{dct} and AugMix~\citep{Hendrycks2020augmix} when training the score network. The idea is, instead of the typical Gaussian distribution, we can use the modified perturbation distribution $q(\tilde{x}|x) = \mathcal{N}(\tilde{x} | F(x), \sigma^2I)$ where $F(x)$ is the augmentation obtained by either \gls{dct} or AugMix. Please refer to \cref{sec:cifar10cfull} for detailed description.

\subsubsection{Evaluation: Extra datasets}
\label{sec:extra}
We also present our results for other image benchmarks, including MNIST, FashionMNIST, and CIFAR-100. For CIFAR-100,  We provide additional results of extensive datasets in~\cref{sec:extensive}. For CIFAR-100, ours achieved remarkable robust accuracy of \textbf{39.72\%} against \gls{bpda} attacks, outperforming the previous result (\citet{Li2020} (adversarial training method) reported the robust accuracy \textbf{28.88\%}) by a large margin.

\subsubsection{Detecting adversarial examples}
Finally, we show that we can detect adversarial examples using the norm of the score networks. \cref{fig:scorehistogram} shows the difference of distributions of the score values for natural and adversarial images.

\section{Conclusion}
\label{sec:conclusion}
In this paper, we proposed a novel adversarial purification method with score-based generative models.
We discovered that an \gls{ebm} trained with \gls{dsm} can quickly purify attacked images with deterministic short-run updates,
and the purification process can further be robustified by injecting Gaussian noises before purification. We validated our method 
on various benchmark datasets using diverse types of adversarial attacks and demonstrated its superior performance.

\section*{Acknowledgement}
This work was supported by Institute of Information \& communications Technology Planning \& Evaluation (IITP) grant funded by the Korea government (MSIT)  (No.2019-0-00075, Artificial Intelligence Graduate School Program (KAIST)). SJH is supported by Institute of Information \& communications Technology Planning \& Evaluation (IITP) grant funded by the Korea government (MSIT) (No.2020-0-00153).
\bibliography{references}
\bibliographystyle{icml2021}
\clearpage
\onecolumn
\icmltitle{Supplementary Materials for Adversarial Purification with Score-based Generative Models}
\vskip 0.3in
\appendix
\section{Experimental details}
\subsection{Software and Hardware Configurations}
We implemented our code on Python version 3.8.5 and PyTorch version 1.7.1 with Ubuntu 18.04 operating system. We run each of our experiments on a single Titan X GPU with Intel Xeon CPU E5-2640 v4 @ 2.40GHz. Our implementation is available at {\tt \href{https://github.com/jmyoon1/adp}{https://github.com/jmyoon1/adp}}.

\subsection{Dataset details}\label{sec:datasets}
\textbf{MNIST} is the dataset that consists of handwritten digits. It consists of a training set of 60,000 examples, and a test set of 10,000 examples. MNIST is a grayscaled dataset with $28\times 28$ size at a total of 784 dimensions, and its label consists of 10 digits. 

\textbf{FashionMNIST} is the dataset that consists of clothes. It consists of a training set of 60,000 examples, and a test set of 10,000 examples. Like MNIST, FashionMNIST is a grayscaled dataset with $28\times 28$ size, where its label is included in one of 10 classes of clothes. The full list of classes is as follows: \{T-shirt/top, Trouser, Pullover, Dress, Coat, Sandal, Shirt, Sneaker, Bag, Ankle boot\}.

\textbf{CIFAR-10} is the dataset that consists of colored images. It consists of a training set of 45,000 examples, a validation set of 5,000 examples, and a test set of 10,000 examples.
CIFAR-10 is an RGB-colored dataset with $32\times32$ size, at a total of 3,072 dimensions each data, where its label belongs to one of the following ten classes. The full list of classes is as follows: \{airplanes, cars, birds, cats, deer, dogs, frogs, horses, ships, trucks\}.

\textbf{CIFAR-100} is also the dataset that consists of colored images.
It consists of a training set of 45,000 examples, a validation set of 5,000 examples, and a test set of 10,000 examples.
Like CIFAR-10, CIFAR-100 is an RGB-colored dataset with $32\times32$ size.

\textbf{CIFAR-10-C} is the dataset that consists of corrupted CIFAR-10 examples.
It consists of 15 types of adversaries, denoted to \textit{common corruption}, with 5 severities each.
We introduce some samples from common corruption examples at \cref{sec:ccexample}.


\subsection{Training hyperparameters}\label{sec:hyperparameter}
\begin{table}[!ht]
\caption{\label{table:hp}Hyperparameters for training score networks.}
\centering
\begin{tabular}{llllll}
\toprule
Dataset & $\sigma_1$ & $\sigma_L$ & $L$ & Training iterations & Batch size \\
\midrule
MNIST & 15 & 0.005253 & 110 & 200,000 & 128  \\
FashionMNIST & 15 & 0.005253 & 64 & 200,000 &128\\
\midrule
CIFAR-10 & 50 & 0.008454 & 232 & 300,000 & 128\\
CIFAR-100 & 50 & 0.008454 & 232 & 300,000 & 128\\
CIFAR-10, \gls{dct} Augmented & 50 & 0.08454 & 232 & 200,000 & 128\\
CIFAR-10, AugMix Augmented & 50 & 0.08454 & 232 & 200,000 & 128\\
\bottomrule
\end{tabular}
\end{table}
We present the hyperparameters that are used for training our purifier networks having \gls{ncsn}v2 architecture in~\cref{table:hp}.
Here, $\sigma_1$ and $\sigma_L$ stands for the largest and smallest standard deviation of the isotropic Gaussian noise for training \gls{ncsn}v2, $L$ is the number of steps of noise standard deviations.
We follow~\citet{Song2020NCSNv2} to get appropriate hyperparameters. When we train \gls{ncsn} with \gls{dct}- or AugMix-augmented perturbations to enhance robustness in CIFAR-10-C evaluation, the smallest noise level $\sigma_L$ is adjusted since out-of-distribution examples might be over-represented for training with small noise levels, because the distance between the original and perturbed inputs will become farther compared to \gls{ncsn} trained with Gaussian perturbations.

For training all the classifier and purifier networks, we use Adam optimizer with learning rate 0.001 and $(\beta_1, \beta_2)=(0.9, 0.999)$, and no weight decay. We disabled horizontal flip at MNIST and FashionMNIST datasets, and enabled it at CIFAR-10, CIFAR-100 datasets.

\subsection{Neural Network Descriptions}\label{sec:nndescription}
\begin{figure}
\centering
    \begin{subfigure}{\textwidth}
    \centering
    \includegraphics[width=.7\linewidth]{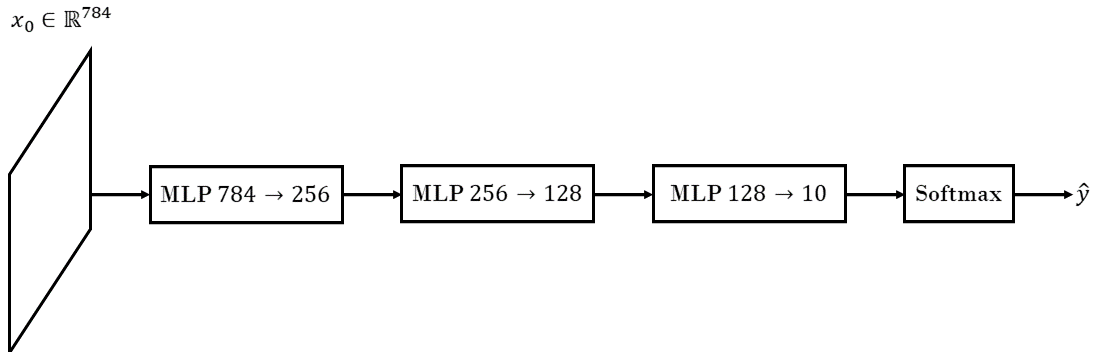}
    \caption{Simple MLP structure for MNIST classifier}
    \end{subfigure}
    \begin{subfigure}{\textwidth}
    \centering
    \includegraphics[width=.7\linewidth]{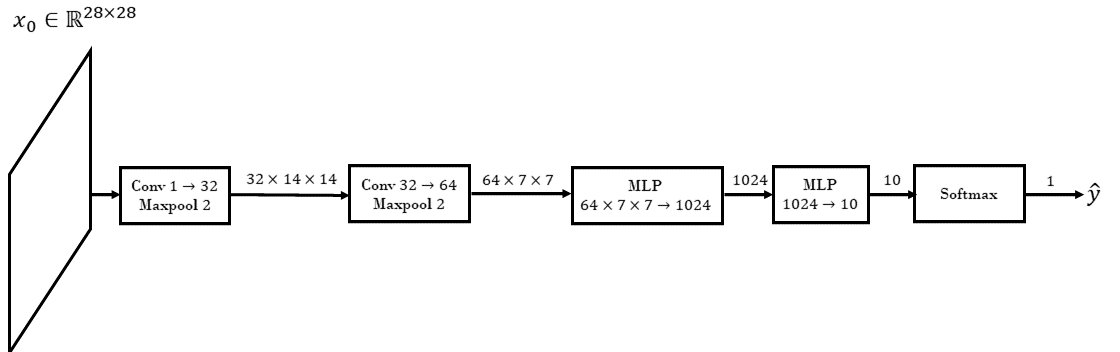}
    \caption{Simple CNN structure for FashionMNIST classifier}
    \end{subfigure}
    \begin{subfigure}{\textwidth}
    \centering
    \includegraphics[width=.9\linewidth]{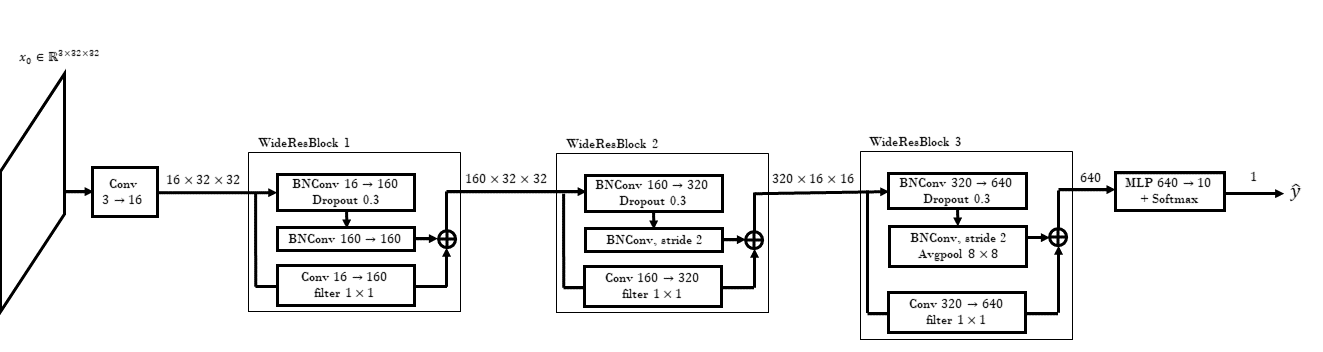}
    \caption{Simple CNN structure for classifying CIFAR-10 and CIFAR-100. For TinyImageNet, all image sizes and avgpool size are doubled.}
    \end{subfigure}
\caption{Neural network architecture for classifier networks}\label{fig:classifier}
\centering
    \begin{subfigure}{\textwidth}
    \centering
    \includegraphics[width=.99\linewidth]{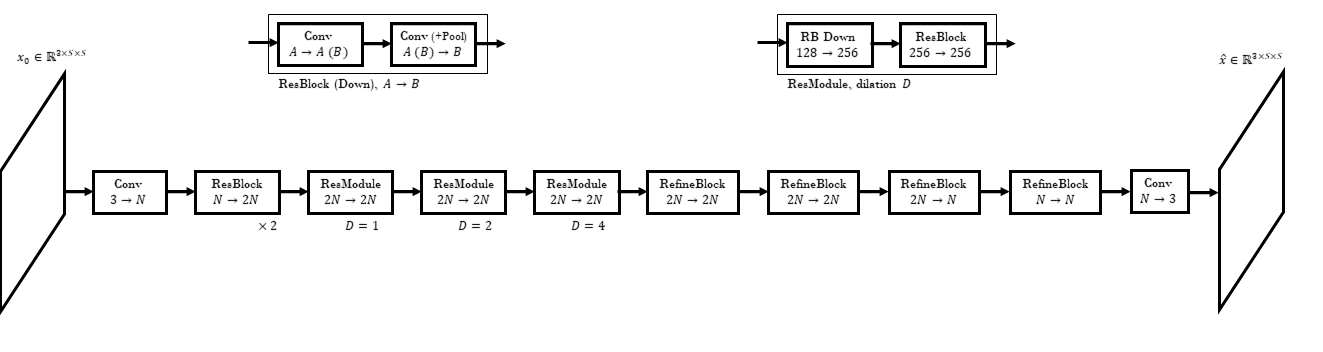}
    \end{subfigure}
\caption{Neural network architecture for the \gls{ncsn} purifier network}\label{fig:purifier}
\end{figure}

For CIFAR-10, CIFAR-10-C and CIFAR-100 datasets, we use WideResNet-28-10~\citep{Zagoruyko2016WRN} for classification and \gls{ncsn}v2~\citep{Song2020NCSNv2} which is a modified version of RefineNet~\citep{Lin2017RefineNet} for purification. The overall structures are depicted in \cref{fig:classifier}. In CNN architecture for FashionMNIST classifier, we use filter size $5\times5$, stride $1$, and padding $2$ in our pytorch implementation. For WideResNet-28-10 architecture for classifier for larger datasets (CIFAR-10, CIFAR-100), we use filter size $3\times3$, stride $1$, and padding $1$ in our Pytorch implementation.

We also describe the detailed \gls{ncsn} structure in \cref{fig:purifier}. Here, $N$ denotes the number of channels. The RefineNet \citep{Lin2017RefineNet} structure is used as the decoder part of \gls{ncsn}.

\begin{figure}[!htb]
\centering
    \begin{subfigure}{\textwidth}
    \centering
    \includegraphics[width=.9\linewidth]{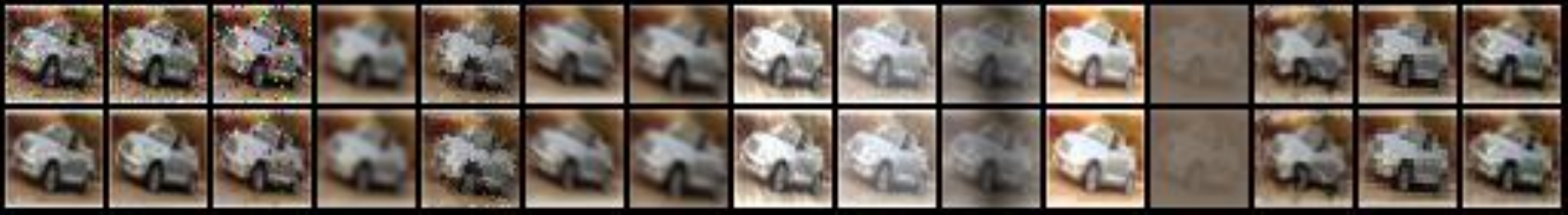}
    \end{subfigure}
\caption{Examples of corrupted and purified images. From left: \{Gaussian, shot, impulse\} noise, \{Defocus, glass, motion, zoom\} blur, \{snow, frost, fog, brightness\} weather, \{contrast, elastic, pixelate, JPEG\} digital corruptions.}\label{fig:corrupted}
\end{figure}

\section{Common corruption and purified examples}
\label{sec:ccexample}

\cref{fig:corrupted} shows the examples of images corrupted with severity level 5 and their corresponding purified counterparts. The order of corrupted images is the same as indicated in \cref{table:cifar10c}.

\section{Full CIFAR-10-C performance}\label{sec:cifar10cfull}
In this section, we report our full common corruption performances.
For implementation of adversarial training methods, we import from the \tt RobustBench\rm~\citep{croce2020robustbench} benchmark.
We present both the full and average performances for every 15 kind of corruptions in \cref{table:cifar10c}.
We used adaptive step size $\alpha=0.05$ for \gls{adp} trained with Gaussian and \gls{dct} perturbation, and exponentially decreasing deterministic step size from $\sigma_1=0.08$ to $\sigma_{10}=8.0\times10^{-4}$ for \gls{adp} trained with AugMix augmented perturbation.
For (\gls{dct}+AugMix) case, we iteratively purify the input with those two purifier models 10 times in rotation.
For comparison, the results at the second category include recent adversarial training cases, and the third category include recent data augmentation and domain adaptation methods.

When training \gls{ncsn} with \gls{dct} or AugMix augmentations, we slightly modified the \gls{dsm} objective.
The idea is, as described in the main text, to modify the perturbation distribution from Gaussian distributions centered at original images to Gaussian distributions centered at augmented images.

For \gls{dct} training of the purifier network, we first take \gls{dct} to original images, and drop the frequency components with smallest eigenvalues, until the sum of dropped coefficients reach 5\% of the sum of their eigenvalues.
Then the smallest noise level $\sigma_L$ in the \gls{ncsn} objective is multiplied by 10, since \gls{dct}-transformed images are more deviated from original images than conventional noisy images and norm-based attacked images in terms of $l_2$-distance, and too small noise level may over-represent the deviation by \gls{dct} transformation.
Then, we replace the perturbation distribution by $q(\tilde{x}|x)=\mathcal{N}(\tilde{x}|F(x),\sigma^2 I)$ where $F(x)$ is a \gls{dct}-transformed image from $x$.

AugMix \cite{Hendrycks2020augmix} has even larger deviation than \gls{dct} augmentation, so it is more difficult to train the purifier network with it.
Instead of directly targeting the original image, we first generate the auxiliary image that locates comparatively near to the perturbed point, then target to the auxiliary image.
We first replace the perturbation distribution by $q'(\tilde{x}|x)=\mathcal{N}(\tilde{x}|F(x),\sigma^2 I)$ where $F(x)$ is an AugMix-transformed image from $x$.
Then we replace the \gls{dsm} objective \cref{eq:dsm_objective} with
\begin{equation}
    \ell(\theta,\sigma) = \mathbb{E}_{q'(\tilde x|x) p_\text{data}(x)}\left[
    \frac{1}{2\sigma^4} \norm{
    \tilde x + \sigma^2s_\theta(x') - x'
    } ^2
    \right]
\end{equation}
where $x'=\frac{x+F(x)}{2}$ is the midpoint of $x$ and $F(x)$.
That is, to ease the reconstruction from highly corrupted images $F(x)$, we choose to learn $s_\theta(x)$ to recover from the midpoint $x'$.

\section{Detailed results for strong adaptive attacks}
\subsection{Full list of defense results for adaptive attacks}
\label{sec:adaptive}
\begin{table}
\caption{\label{table:adaptivefull}Evaluation results for adaptive attacks. Threat model: $l_\infty$ $\varepsilon$-ball with $\varepsilon=8/255$, CIFAR-10 dataset. The white-box attack results for adversarial training methods are also referred for comparison.}
\small
\centering
\begin{tabular}{lllllll}
\toprule
 & Natural & Robust & Preprocessor & Classifier & Attack method&Threat blindness\\
\midrule
\gls{adp} (Adaptive LR) & 86.14 & & & & & \\
\hspace{3mm}($\sigma=0.25$), \gls{bpda} step 40 &  & 70.01 & \gls{ncsn}v2 & WRN-28-10 & \gls{bpda}+\gls{eot} & Unseen \\
\hspace{3mm}($\sigma=0.25$), \gls{bpda} step 100 &  & 69.71 & \gls{ncsn}v2 & WRN-28-10 & \gls{bpda}+\gls{eot} & Unseen \\
\hspace{3mm}($\sigma=0.25$) & & 70.61 & \gls{ncsn}v2 & WRN-28-10 & Joint (score)+\gls{eot} & Unseen \\
\hspace{3mm}($\sigma=0.25$) & & 78.39 & \gls{ncsn}v2 & WRN-28-10 & Joint (full)+\gls{eot} & Unseen \\
\hspace{3mm}($\sigma=0.25$) & & 80.80 & \gls{ncsn}v2 & WRN-28-10 & \gls{spsa} & Unseen \\
\hspace{3mm}($\sigma=0.25$) with detection & 95.74 & 69.85 & \gls{ncsn}v2 & WRN-28-10 & \gls{bpda}+\gls{eot} & Unseen \\
\midrule
\citep{hill2020stochastic} (1500 iterations) & 84.12 & 54.90 & IGEBM & WRN-28-10 & \gls{bpda}+\gls{eot} & Unseen\\
\citep{pmlr-v97-yang19e} (Natural)& 94.8 & 40.8 & Masking+Recon.& ResNet-18 & \gls{bpda} & Unseen\\
\citep{pmlr-v97-yang19e} (AT) & - & 52.8 & Masking+Recon.& ResNet-18 & \gls{bpda} & Unseen\\
\citep{pmlr-v97-yang19e} (AT) & 88.7 & 55.1 & Masking+Recon.& WRN-28-10 & \gls{bpda} & Seen\\
\citep{pmlr-v97-yang19e} (Natural) & 89.4 & 41.5 & Masking+Recon.& ResNet-18 & Approx. Input&Unseen\\
\citep{pmlr-v97-yang19e} (AT) & 88.7 & 62.5 & Masking+Recon. & ResNet-18 & Approx. Input & Seen\\
\citep{song2018pixeldefend} & 95 & 5 & PixelCNN & ResNet-62 & \gls{bpda} & Seen\\
\midrule
\citep{madry2018towards} & 87.3 & 45.8 & Robust Classifier & ResNet-18 & Full \gls{pgd} & Seen\\
\citep{DBLP:conf/icml/ZhangYJXGJ19} & 84.90 & 56.43 & Robust Classifier & ResNet-18 & Full \gls{pgd} & Seen \\
\citep{Carmon2019RST} & 89.70 & 62.50 & Robust Classifier & WRN-28-10 & Full \gls{pgd} & Seen\\
\bottomrule
\end{tabular}
\end{table}
In this section, we present the full list of defense results for strong adaptive attacks contained in~\cref{table:ad} of the main paper in~\cref{table:adaptivefull}, including the preprocessor and classifier architectures, attack method, and blindness against the threat model.
The results at the first, second, and third category includes our work, recently proposed preprocessor-based defense methods, and existing adversarial training-based defense methods, respectively.
\emph{Approx. Input}~\citep{pmlr-v97-yang19e} first iteratively updates inputs by classifier \gls{pgd} followed by purification,
and thus add classifier gradients to purified images instead of clean images. The term \emph{with detection} denotes our method with the procedure of detecting adversarial examples before the purification, as described in \cref{sec:detection}. 
Our method with detection can increase clean accuracy because it can filter out natural images and prevent the need of unnecessary purification.

\subsection{Performance with various noise injection levels}
\label{sec:othersigma}
\begin{table}[t]
\centering
\small
\caption{CIFAR-10 results for different levels of noise injections on attacked images, from $\sigma=0.05$ to $\sigma=0.4$ with preprocessor-blind classifier \gls{pgd} attacks.}
\label{table:diffsigma}
\begin{tabular}{llllll}
\toprule
\multirow{2}{*}{Method} & \multicolumn{3}{c}{Accuracy (\%)} & \multirow{2}{*}{Architecture} & \multirow{2}{*}{Blindness} \\
\cmidrule{2-4}
 & Standard & \gls{bpda}+\gls{eot} & Clf \gls{pgd} & & \\
\midrule
\gls{adp} ($\sigma=0.05$) & 93.35 & 6.08 & 66.94 & WRN-28-10 & Unseen \\  
\gls{adp} ($\sigma=0.10$) & 93.09 & 41.06 & \textbf{87.13} & WRN-28-10 & Unseen \\
\gls{adp} ($\sigma=0.15$) & 90.36 & 57.73 & 86.34 & WRN-28-10 & Unseen \\
\gls{adp} ($\sigma=0.20$) & 86.80 & 67.36 & 85.74 & WRN-28-10 & Unseen \\
\gls{adp} ($\sigma=0.25$) & 86.14 & \textbf{70.01} & 83.93 & WRN-28-10 & Unseen \\
\gls{adp} ($\sigma=0.30$) & 80.98 & 69.06 & 78.89 & WRN-28-10 & Unseen \\
\gls{adp} ($\sigma=0.35$) & 79.44 & 69.70 & 77.54 & WRN-28-10 & Unseen \\
\gls{adp} ($\sigma=0.40$) & 77.41 & 69.67 & 75.80 & WRN-28-10 & Unseen \\
\bottomrule
\end{tabular}
\end{table}
We present the standard and robust accuracy of \gls{adp} for the strong adaptive \gls{bpda}+\gls{eot} attack as well as the preprocessor-blind classifier \gls{pgd} attacks in CIFAR-10 dataset, from $\sigma=0.05$ to $\sigma=0.4$ in~\cref{table:diffsigma}.
As the noise level increases, both the standard accuracy and the gap between standard and robust accuracy decrease, as the standard accuracy falls much faster than the robust accuracy as the injected noise becomes stronger.
Although both attacks are held in the same threat models, the best robust accuracy at the classifier \gls{pgd} attack is achieved at much less injected noise than the \gls{bpda}+\gls{eot} attack, implying that the classifier \gls{pgd} attack actually needs less noise injection than \gls{bpda}+\gls{eot} attack for optimal purification.


\subsection{Effect of number of \gls{eot} for \gls{bpda} attacks}
\label{sec:bpdaeot}
\begin{table}
\caption{\gls{bpda}+\gls{eot} attack results for different number of \gls{eot} in CIFAR-10 dataset. The input is attacked after (\#\gls{eot}) different random noise injections with $\sigma=0.25$ via \gls{bpda} attack.}
\centering
\setlength{\tabcolsep}{3pt}
\begin{tabular}{llllllll}
\toprule
Number of \gls{eot} & 1 & 3 & 5 & 10 & 15 & 30 & 50 \\
\midrule
Accuracy (\%) & 76.46 & 73.89 & 72.90 & 69.86 & 70.01 & 68.75 & 67.60 \\
\bottomrule
\end{tabular}
\label{table:bpdaeot}
\end{table}
We present the robust accuracy of \gls{adp} over \gls{bpda}+\gls{eot} attacks with different number of \gls{eot} in~\cref{table:bpdaeot} in CIFAR-10 dataset.

\subsection{Effect of \gls{bpda} iterations}
The experiments of the main paper are all performed with 40 iterations of \gls{bpda} attacks. To measure the effect of the number of \gls{bpda} iterations, we also run experiments with 100 iterations of \gls{bpda} attacks. As contained in~\cref{table:adaptivefull}, increasing the number of iterations from $40$ to $100$ slightly decrease the robust accuracy by 0.30\%.

\subsection{Effect of purification runs}
\begin{figure}
    \centering
    \includegraphics[width=.5\linewidth]{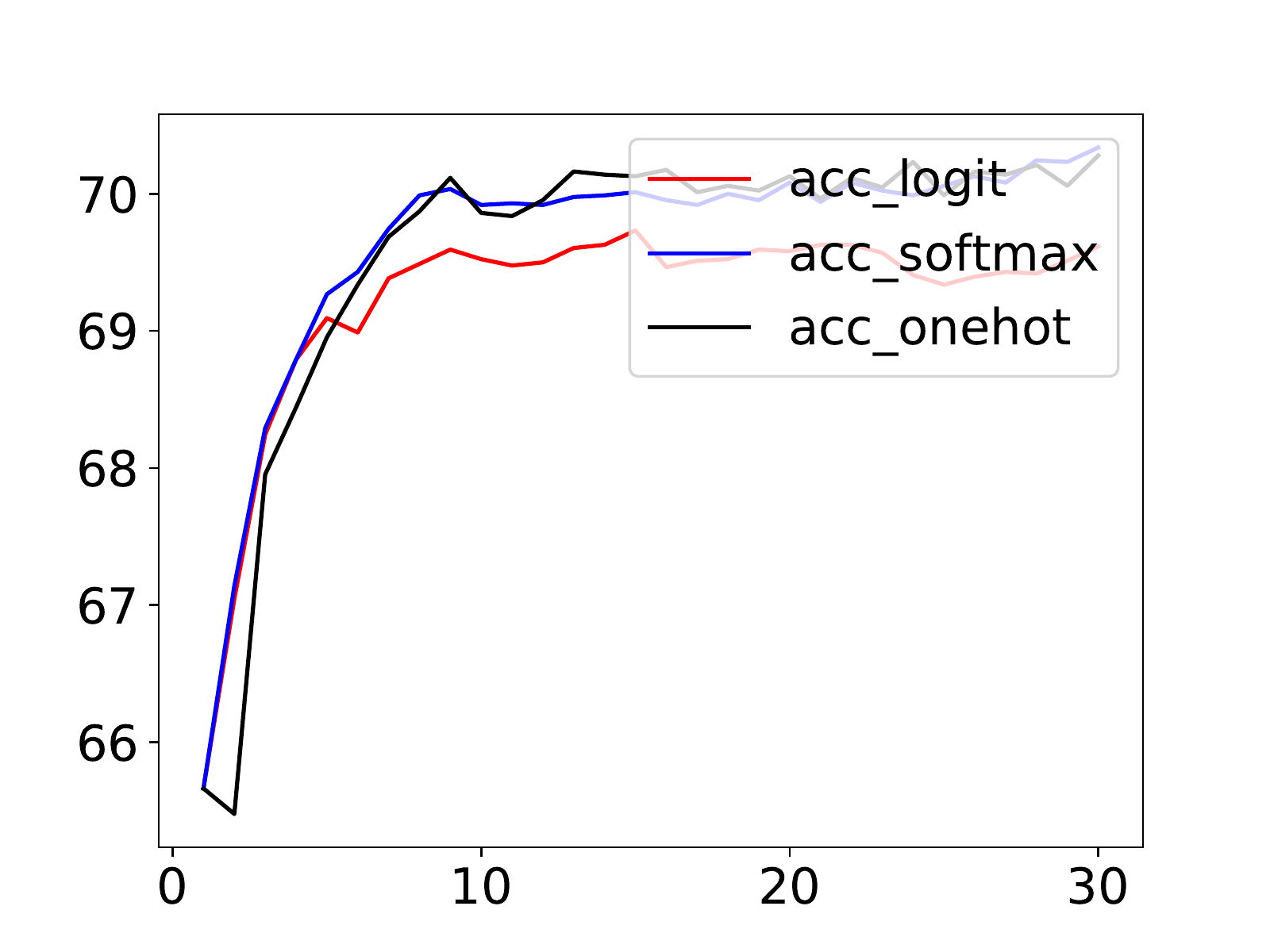}
    \caption{Robust accuracy of CIFAR-10 dataset under \gls{bpda}+\gls{eot} attack on different purification runs. The x-axis stands for the number of purification runs and y-axis stands for accuracy (\%). The {\color{red}red}, {\color{blue}}, and black line stand for expectation over {\color{red}pre-softmax outputs}, {\color{blue}post-softmax outputs}, and argmax outputs, respectively.}
    \label{fig:purrun}
\end{figure}

In the main paper, we fixed the maximum number of purification runs to 10. In this section, we present robust accuracy of \gls{adp} on different number of purification runs in CIFAR-10 dataset, with \gls{bpda}+\gls{eot} attack with 40 \gls{bpda} iterations and 15 \gls{eot} attacks. As described in \cref{fig:purrun}, we observed that the robust accuracy is improved until 10 runs, and stay stable for more runs.

\subsection{Full list of defense results for more datasets}
\label{sec:extensive}
We present the full list of defense results for various datasets, including MNIST, FashionMNIST, and CIFAR-100 in~\cref{table:extensive}.

\section{Robust accuracy of Randomized Smoothing Classifiers}
\label{sec:randsmoothing}
We present the standard accuracy of randomized smoothing classifiers of \gls{adp} on CIFAR-10 dataset in~\cref{table:certified}. We see that on low noise levels up to $\sigma=0.25$, the robust accuracy of the randomized smoothing classifier performing \gls{adp} surpasses those of the existing randomized smoothing classifiers.
\begin{table}[!htb]
\centering
\caption{Evaluation results for more datasets.}
\begin{tabular}{llllll}
\toprule
Dataset & \multirow{2}{*}{$\varepsilon$} & \multirow{2}{*}{Attack type} & \multicolumn{2}{c}{Accuracy (\%)} \\
\hspace{3mm}Defense methods & & & Standard & Robust \\
\midrule
MNIST & 0.3 & Clf \gls{pgd} & 98.07 & 96.41 \\ 
FashionMNIST & $8/255$   & Clf \gls{pgd} & 93.19 & 86.62 \\ 
\midrule
CIFAR-100 & & & & \\
Raw WideResNet & & & 79.86 & \\
\hspace{3mm}$\sigma=0.0$, det 0.08 & $8/255$  & Clf \gls{pgd} & 77.83 & 43.21 \\ 
\hspace{3mm}$\sigma=0.25$, $\alpha=0.05$ & $8/255$ & \gls{bpda}+\gls{eot} & 60.66 & 39.72 \\
\citep{hill2020stochastic} & $8/255$ & \gls{bpda}+\gls{eot} & 51.66 & 26.10 \\
AT~\citep{madry2018towards} & $8/255$ & \gls{pgd} & 59.58 & 25.47 \\
\citep{Li2020} & $8/255$ & \gls{pgd} & 61.01 & 28.88 \\
\bottomrule
\end{tabular}
\label{table:extensive}
\end{table}
\begin{table}
\caption{Robust accuracy of randomized smoothing classifiers.}
\centering
\begin{tabular}{lllll}
\toprule
\multirow{2}{*}{Models} & \multicolumn{4}{c}{Noise level $\sigma$}  \\
 & 0.12 & 0.25 & 0.5 & 1.0 \\
\midrule
\gls{adp} & {93} & {86} & 62 & 27 \\
\cite{Cohen2019randomized}* & 81 & 75 & 65 & 47 \\
\cite{Salman2019Provably}* & 84 & 77 & {68} & {50} \\
\bottomrule
\end{tabular}
\label{table:certified}
\end{table}

\section{Detecting Adversarial Examples before purification}
\label{sec:detection}
\begin{figure}
    \begin{subfigure}[b]{.33\textwidth}
      \centering
      \includegraphics[width=\linewidth]{./figures/hist_score_pgd.pdf}
      \caption{}
      \label{fig:histscorepgd}
    \end{subfigure}
    \begin{subfigure}[b]{.33\textwidth}
      \centering
      \includegraphics[width=\linewidth]{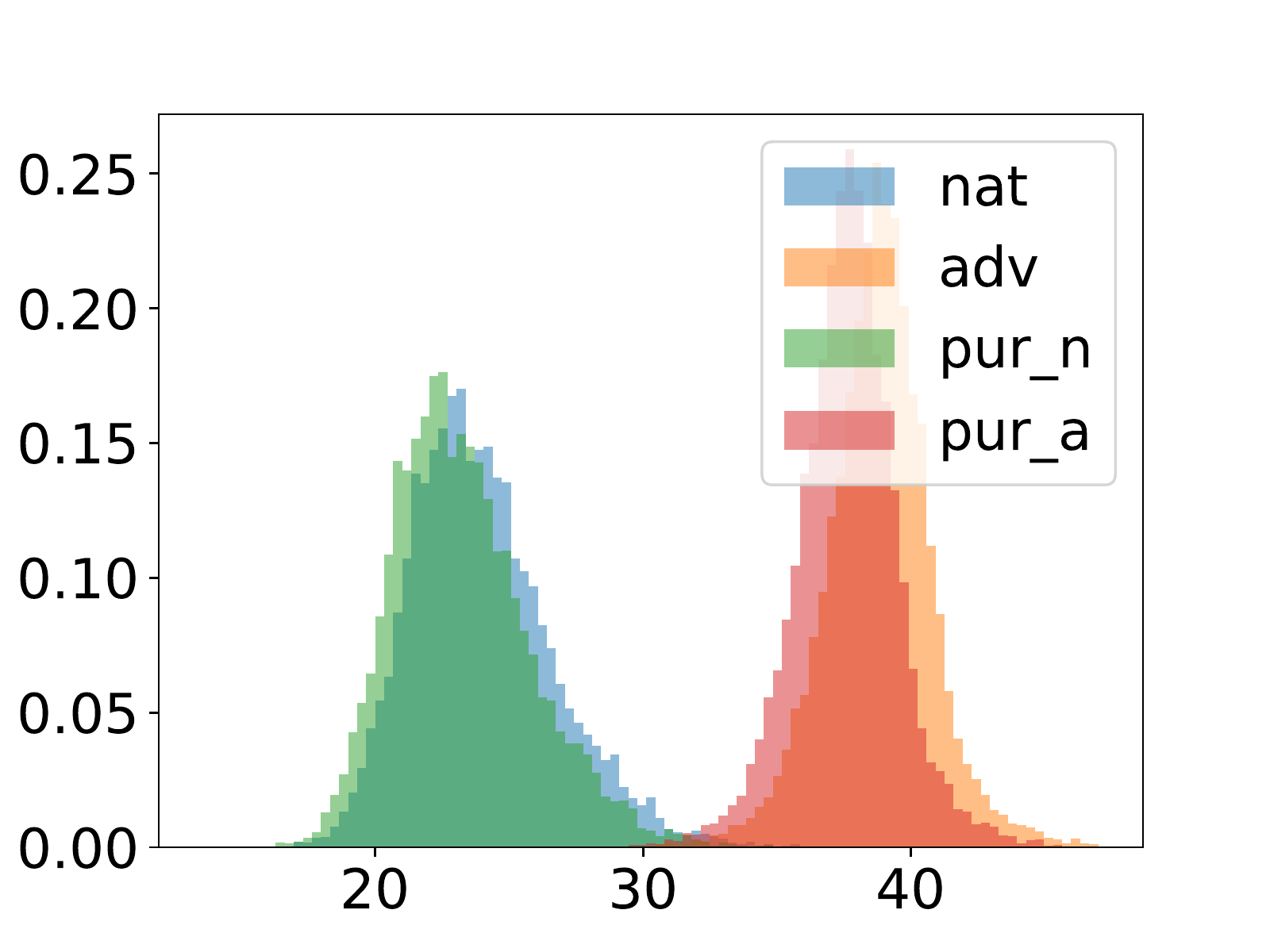}
      \caption{}
      \label{fig:histscorebpda}
    \end{subfigure}
    \begin{subfigure}[b]{.33\textwidth}
      \centering
      \includegraphics[width=\linewidth]{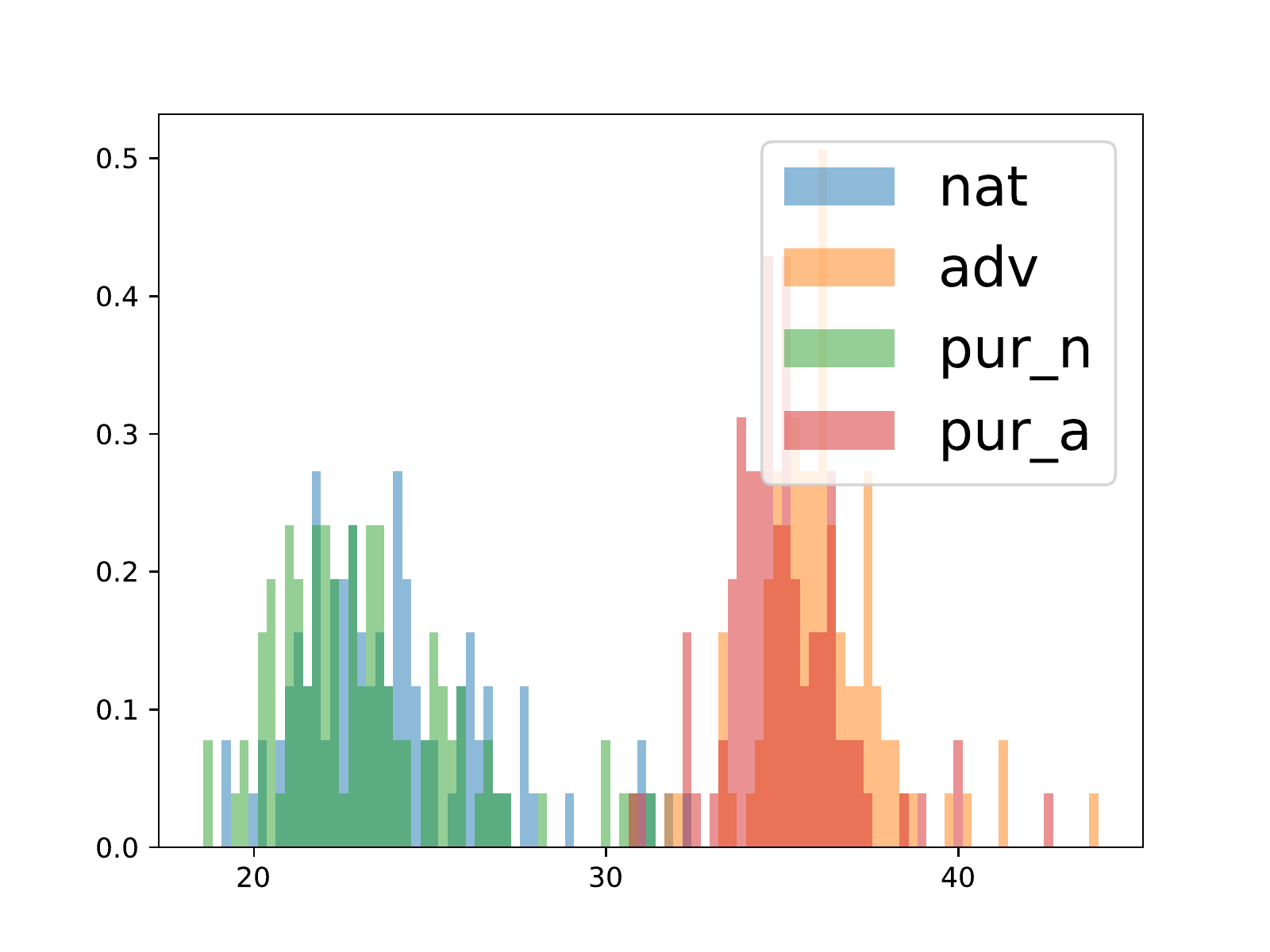}
      \caption{}
      \label{fig:histscoreapprox}
    \end{subfigure}
    \begin{subfigure}[b]{.33\textwidth}
      \centering
      \includegraphics[width=\linewidth]{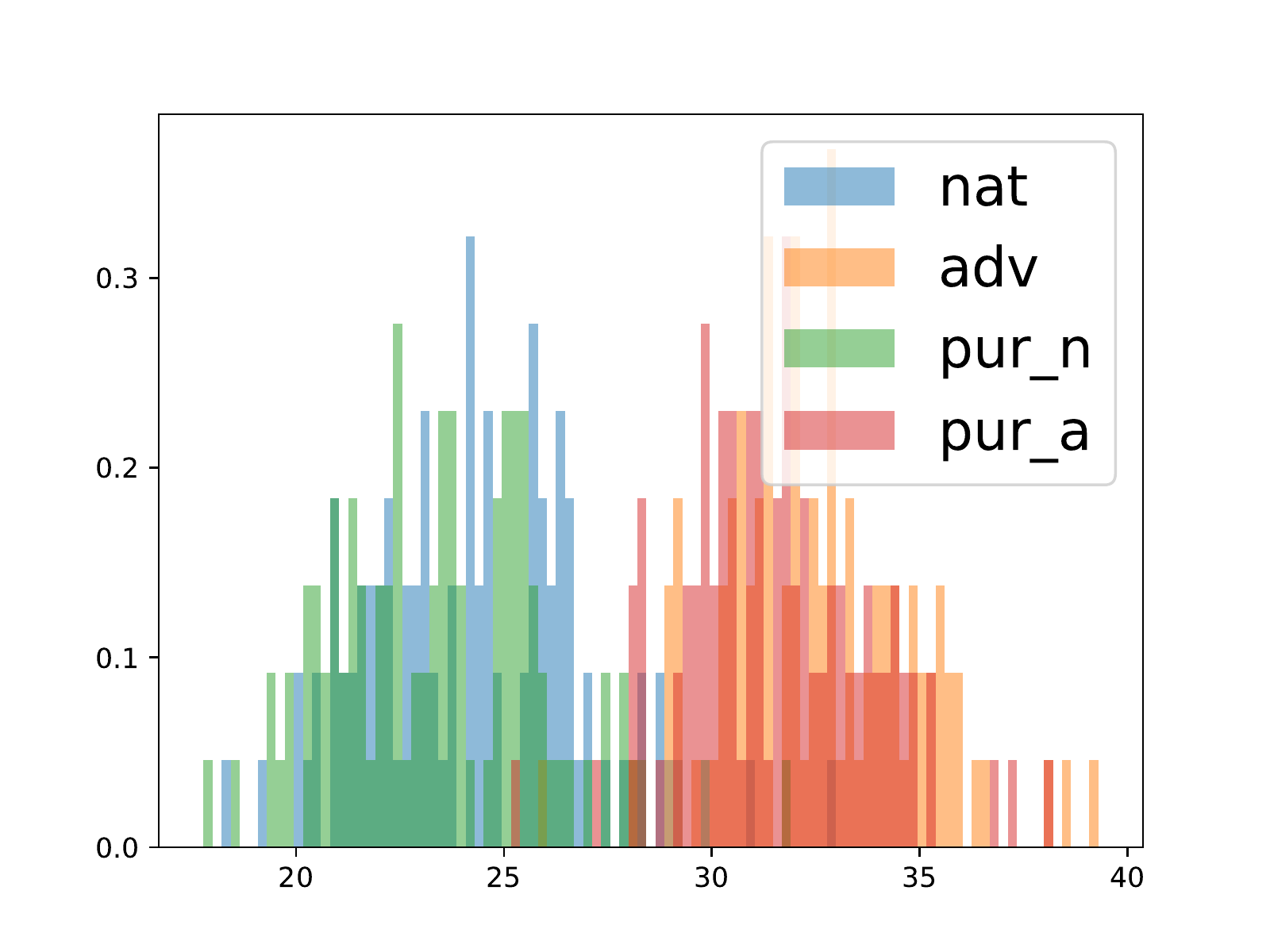}
      \caption{}
      \label{fig:histscoreunrolling}
    \end{subfigure}
    \begin{subfigure}[b]{.33\textwidth}
      \centering
      \includegraphics[width=\linewidth]{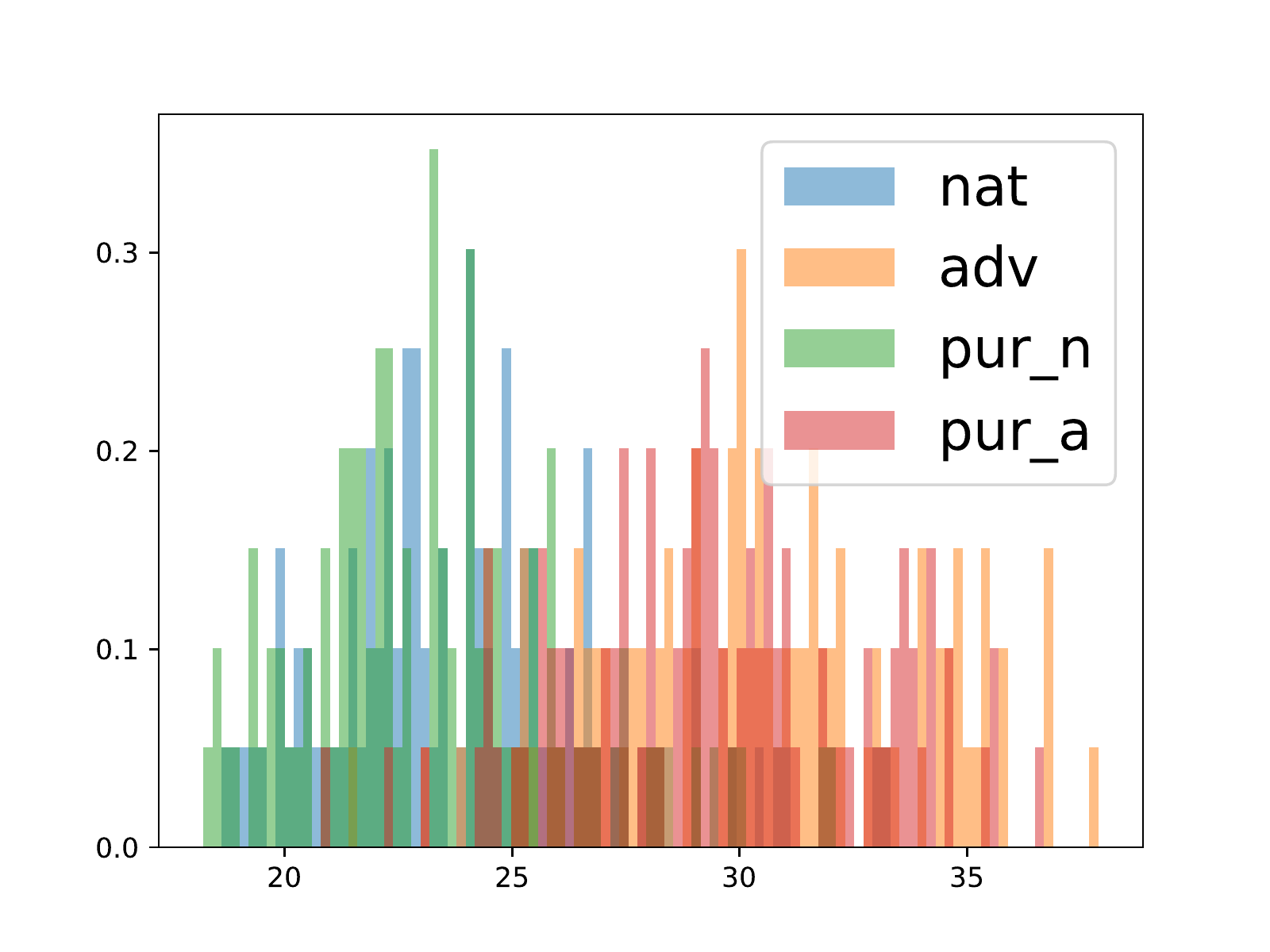}
      \caption{}
      \label{fig:histscorejointfull}
    \end{subfigure}
    \begin{subfigure}[b]{.33\textwidth}
      \centering
      \includegraphics[width=\linewidth]{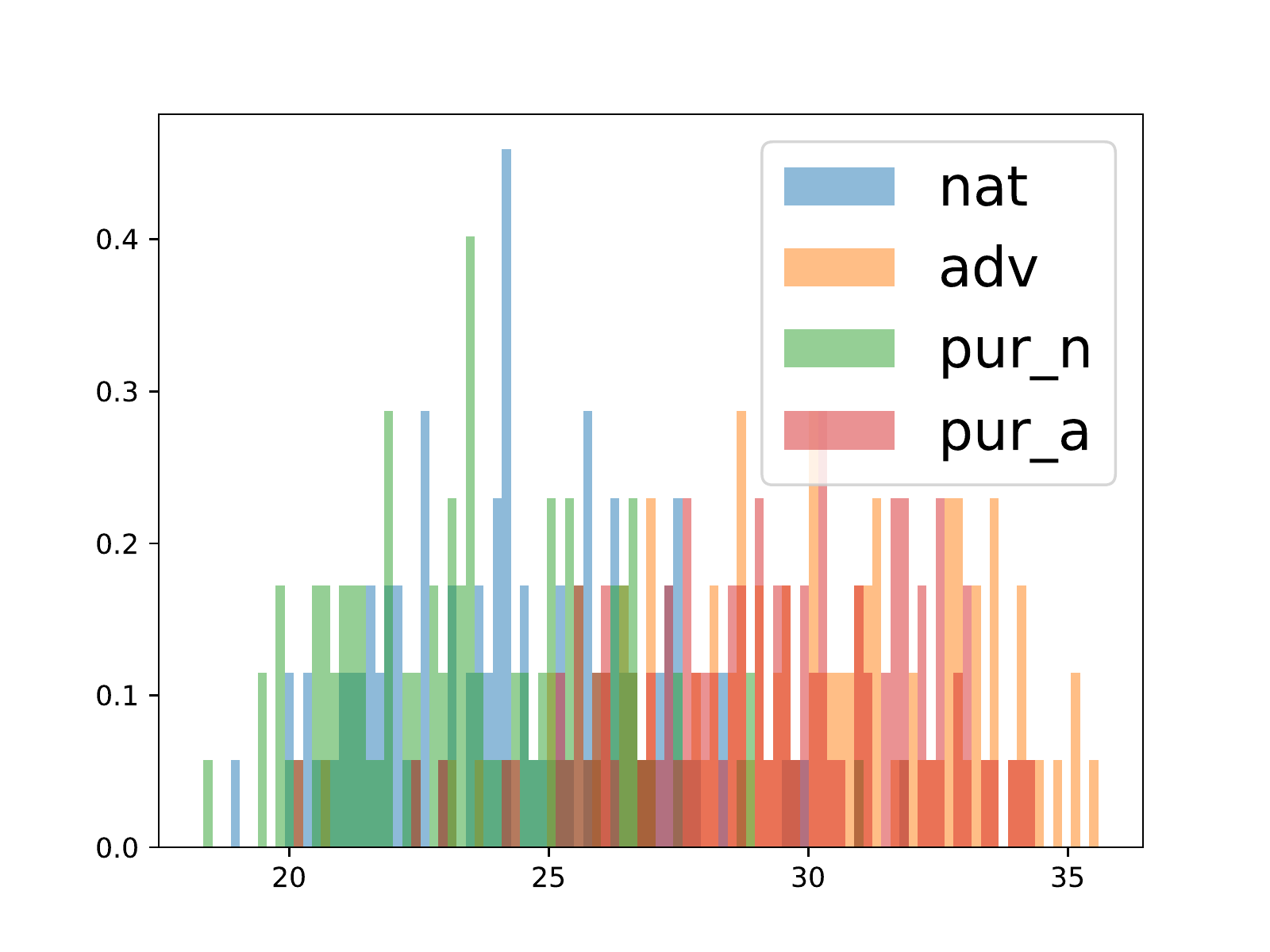}
      \caption{}
      \label{fig:histscorejointone}
    \end{subfigure}
    \caption{Histogram of score function norms $\norm{s_{\theta}(x)}$ for natural, adversarial and purified images. Pur\_a and Pur\_n denotes scores of one-step purified adversarial and natural images, respectively.
    The x-axis and y-axis stand for the score norm $\norm{s_{\theta}(x)}_2$ and the probability density, respectively.
    From upper left to lower right: (a) Classifier \gls{pgd} (b) \gls{bpda} (c) Approximate input (d) One-step unrolling (e) Joint (full) (f) Joint (score).
    The x-axis and y-axis represent $\norm{s_{\theta}(x)}$ and the probability density, respectively. One step unrolling attack is an adaptive attack where the \gls{pgd} attack is performed under the composition of the classifier and one-step forward propagation of the purifier network.}
    \label{fig:detectadv}
\end{figure}
While random noise injection before purification improves the robust accuracy, this degrades the standard accuracy because the features helpful for natural image classification can also be screened out.
To prevent this, we propose a detection and noise injection scheme where we first classify an image into attacked or natural image and apply different noise injection policies according to the classification result.
We draw the histogram of the score norms $\norm{s_{\theta}(x)}$ for natural, adversarial and purified images for various attacks in~\cref{fig:detectadv}.
Except for joint attacks, attacked images usually have higher score norms than natural images, showing the promises of our method for detecting adversarial examples before purifications.

The detection of the attacked is based on the score norms. We choose the threshold $\tau$, and classify an image whose Euclidean norm of the estimate score $\norm{s_{\theta}(x)}_2$ below the threshold as a natural image, and an image whose score norm above the threshold as an attacked image.
\Cref{fig:scorehistogram} shows the histograms of score norms for natural and attacked images. As shown in the figure, the score norm is a good criterion for detecting adversarial examples.

Having decided that an image is an attacked image, we inject higher noise level $\sigma_{\textrm{high}}=\sigma$ (the one obtained with the heuristic described in \cref{sec:methods}).
Otherwise, we apply the low noise level $\sigma_{\textrm{low}} = \beta\sigma$ with $\beta$ fixed to $0.2$. For all experiments on CIFAR-10 and CIFAR-100 datasets, we fixed $\tau=25.0$. 


\section{Decision Boundary Plot with $t$-SNE}
\begin{figure}
\centering
\begin{subfigure}{.33\textwidth}
  \centering
  \includegraphics[width=.99\linewidth]{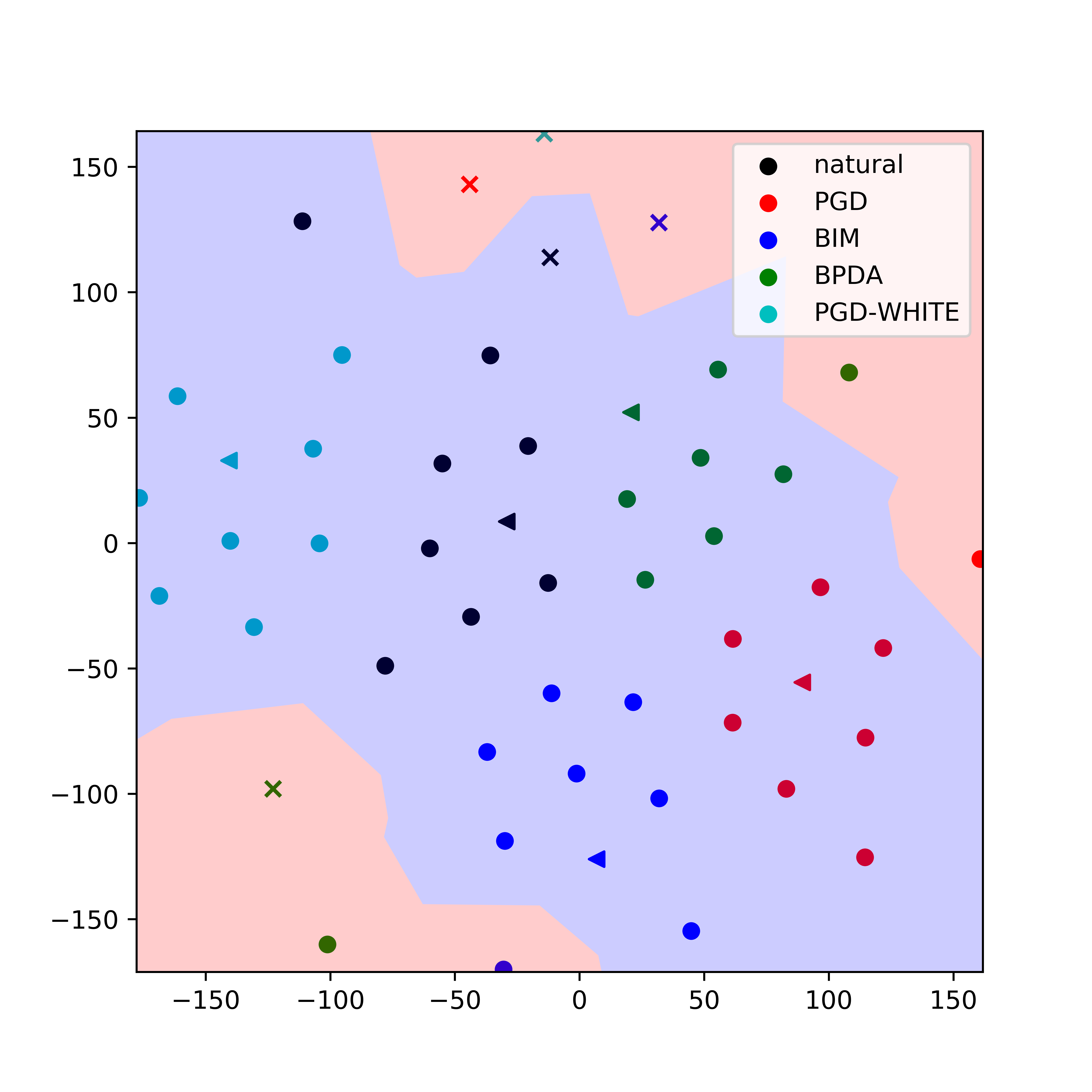}
  \caption{}
\end{subfigure}%
\begin{subfigure}{.33\textwidth}
  \centering
  \includegraphics[width=.99\linewidth]{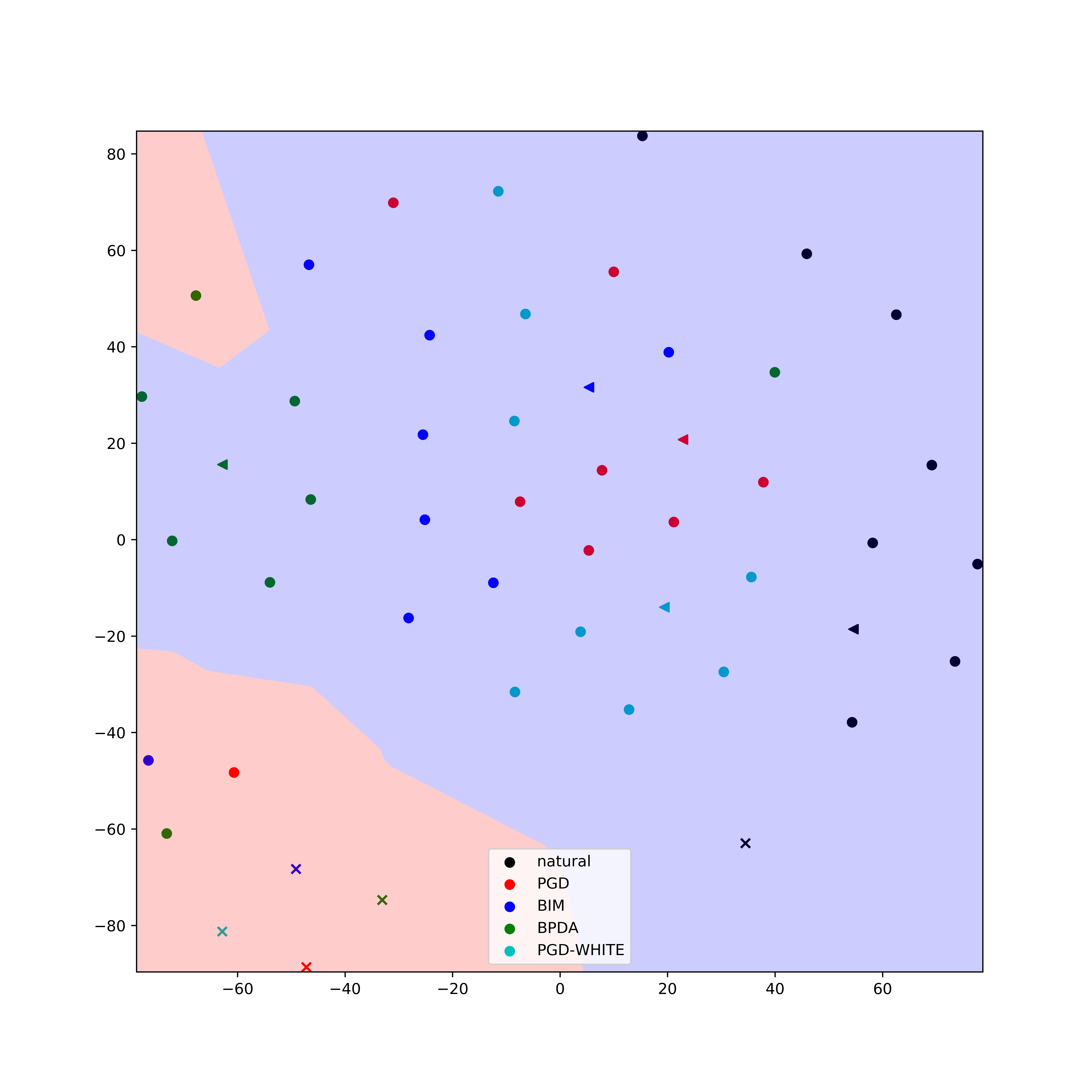}
  \caption{}
\end{subfigure}
\begin{subfigure}{.33\textwidth}
  \centering
  \includegraphics[width=.99\linewidth]{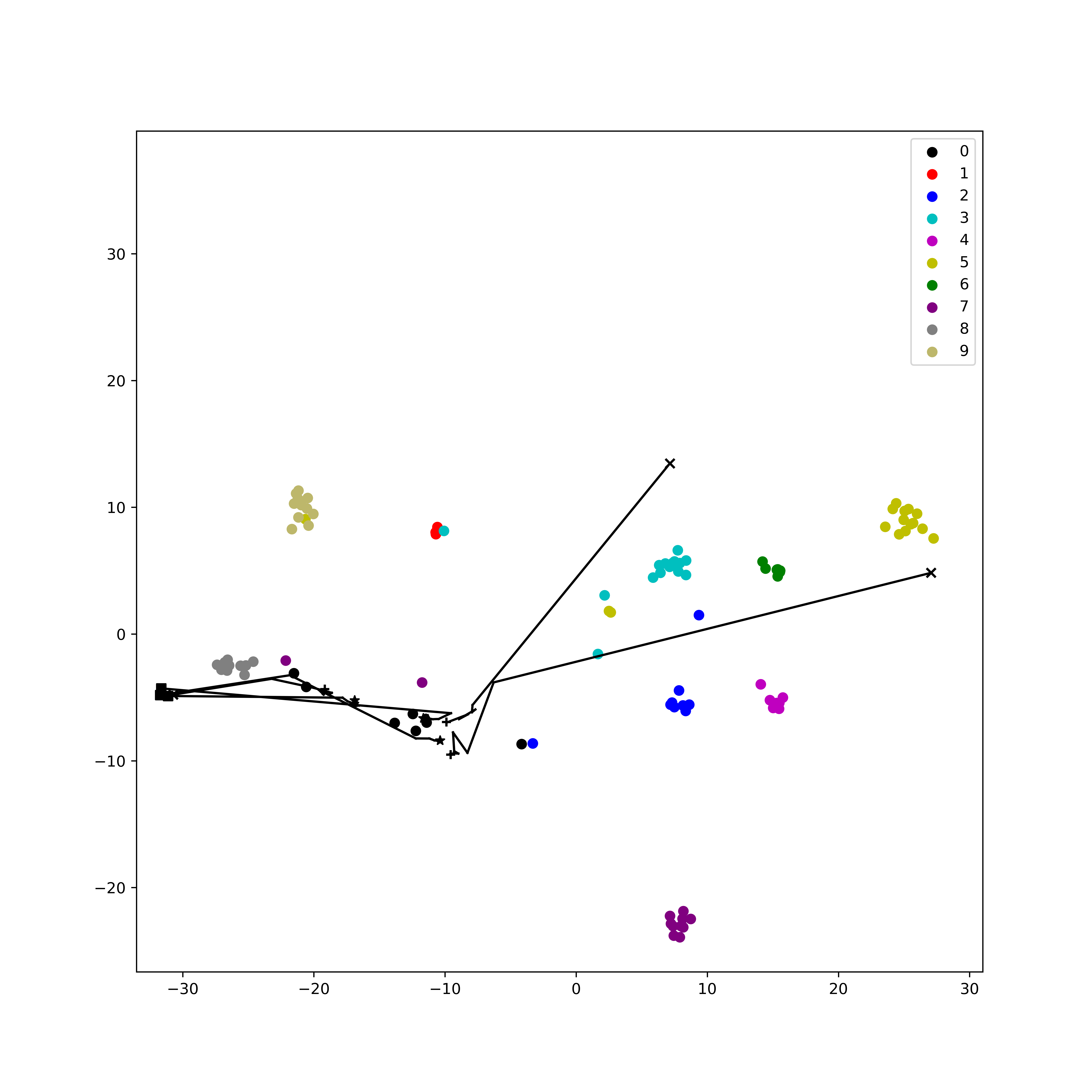}
  \caption{}
\end{subfigure}
\caption{t-SNE diagram of decision boundaries on (a) Image space and (b) Feature space, with respect to various attacks on a single data. The blue and red regions represent the region whose predictions are equal to and different to the ground truth labels, respectively. (c) shows the trajectory of features of attacked images to predicted ones.}\label{fig:moretsne}
\end{figure}

\cref{fig:moretsne} shows the decision boundaries and trajectories over purification steps for existing attacks. We draw $t$-SNE~\citep{vanDerMaaten2008} diagrams for attacked and purified images and their corresponding features, and draw Voronoi diagrams to discriminate between correctly classified images and failed ones. Moreover, we display a trajectory of purifying image drawn on the $t$-SNE diagrams, starting from the attacked images and ends with the purified images. We show that the features of attacked images locate far from the natural images in the feature domain, and approaches to those of the natural images via the purification process.

\clearpage
\begin{landscape}
\begin{table}
\setlength{\tabcolsep}{3pt}
\scriptsize
\caption{\label{table:cifar10c}Performance for CIFAR-10-C dataset}
\vspace{3pt}
\centering
\begin{tabular}{lllllllllllllllll}
\toprule
\multirow{2}{*}{Models} & & \multicolumn{3}{c}{Noise} & \multicolumn{4}{c}{Blur} & \multicolumn{4}{c}{Weather} & \multicolumn{4}{c}{Digital} \\
\cmidrule(lr){3-5}\cmidrule(lr){6-9}\cmidrule(lr){10-13}\cmidrule(lr){14-17}
 & Average & Gaussian & Shot & Impulse & Defocus & Glass & Motion & Zoom & Snow & Frost & Fog & Brightness & Contrast & Elastic & Pixelate & JPEG \\
 \midrule
Raw WideResNet & 71.89\hspace{4mm} & 32.94\hspace{4mm} & 47.78\hspace{4mm} & 46.37\hspace{4mm} & 82.67\hspace{4mm} & 50.41\hspace{4mm} & 78.51\hspace{4mm} & 79.38\hspace{4mm} & 83.86\hspace{4mm} & 78.71\hspace{4mm} & 87.94\hspace{4mm} & 94.43\hspace{4mm} & 75.21\hspace{4mm} & 84.84\hspace{4mm} & 77.52\hspace{4mm} & 77.76\hspace{4mm} \\
\gls{adp} & & & & & & & & & & & & & & & & \\
\hspace{3mm}$\sigma=0.0$ & 80.49 & 91.43 & 91.07 & 71.26 & 82.55 & 52.36 & 78.36 & 79.20 & 83.99 & 79.26 & 87.85 & 94.44 & 75.07 & 84.96 & 77.76 & 77.74 \\
\hspace{3mm}$\sigma=0.25$ & 77.45 & 84.87 & 85.09 & 84.08 & 81.54 & 66.19 & 76.21 & 78.13 & 83.13 & 79.04 & 62.99 & 86.76 & 48.04 & 79.90 & 82.53 & 83.21 \\
\hspace{3mm}$\sigma=0.25$+Detection & 78.96 & 84.86 & 85.09 & 83.75 & 78.58 & 58.68 & 74.13 & 75.64 & 84.88 & 83.26 & 76.47 & 92.75 & 53.06 & 83.45 & 83.87 & 86.00 \\
\hspace{3mm}$\sigma=0.1$ & 76.25 & 88.80 & 88.52 & 83.32 & 74.82 & 62.56 & 67.18 & 69.64 & 80.70 & 81.48 & 64.08 & 89.26 & 52.20 & 75.72 & 81.86 & 83.64 \\
\hspace{3mm}(+DCT Augmentation) & 67.67 & 82.04 & 83.40 & 79.20 & 63.76 & 56.60 & 53.32 & 59.40 & 76.8 & 75.44 & 50.88 & 81.64 & 40.24 & 63.00 & 74.28 & 75.04 \\
\hspace{3mm}$\sigma=0.0$, deterministic LR & 80.09 & 90.28 & 89.55 & 68.81 & 82.53 & 51.36 & 78.39  & 79.20 & 83.95 & 79.13 & 88.09 & 94.49 & 75.40 & 84.83 & 77.58 & 77.79 \\
\hspace{3mm}(+DCT Augmentation) & 80.74 & 84.94 & 86.79 & 81.24 & 80.29 & 58.67 & 74.53 & 76.64 & 86.13 & 85.55 & 87.58 & 94.25 & 73.87 & 82.72 & 78.86 & 82.49 \\
\hspace{3mm}($\times 10$ training var) & 82.63 & 88.60 &90.32 & 83.64 & 82.36 & 62.00 & 76.64 & 79.80 & 87.68 & 87.00 & 87.64 & 92.96 & 76.40 & 80.68 & 78.68 & 85.04 \\
\hspace{3mm}(DCT+AugMix) & 82.40 & 87.00 & 89.48 & 78.68 & 85.92 & 55.84 & 80.04 & 81.40 & 84.64 & 84.80 & 89.72 & 93.28 & 78.48 & 83.24 & 80.44 & 83.04 \\
\midrule
TRADES \citep{DBLP:conf/icml/ZhangYJXGJ19} & 75.63 & 79.17 & 80.45 & 73.85 & 80.05 & 77.96 & 76.50 & 78.97 & 80.42 & 76.58 & 60.30 & 82.63 & 43.11 & 78.87 & 82.73 & 82.81 \\
RST \citep{Carmon2019RST} & 80.40 & 82.49 & 84.14 & 76.98 & 85.47 & 81.71 & 81.92 & 84.65 & 84.57 & 82.70 & 65.90 & 87.59 & 49.01 & 84.05 & 87.68 & 87.20 \\ 
\citep{Cohen2019randomized} & 73.70 & 82.69 & 82.95 & 78.81 & 74.81 & 74.37 & 69.12 & 72.09 & 76.90 & 74.90 & 56.89 & 80.09 & 45.13 & 73.89 & 80.66 & 82.14 \\
\midrule
AugMix \citep{Hendrycks2020augmix}& 88.78 &	81.68 &	86.52 &	85.78 &	94.21 &	79.35 &	92.23 &	92.94 	&89.69 &	89.37 &	91.73 &	94.24 &	90.14 &	90.31 &	86.06 &	87.40 \\
TENT \citep{wang2021tent}&89.52 &	85.22 &	87.82 &	83.78 &	93.84 & 80.04 & 91.82 & 93.02 & 89.58 & 89.84 &	93.56& 94.54 & 94.10 & 89.54 & 91.50 & 84.66 \\
DCT \citep{hossain2020robust} & 89.17 & 85.10 & 88.90 & 86.40 & 94.60 & 78.60 &	90.20 & 91.60 & 89.30 & 90.40 & 91.20 & 94.10 & 80.70 & 90.50 & 91.70 & 94.20 \\ 
\bottomrule
\end{tabular}
\end{table}
\end{landscape}

\end{document}